\documentclass[12pt,a4paper,roman]{aa}  
\usepackage[english]{babel}
\usepackage{setspace}
\usepackage[utf8]{inputenc}
\usepackage{polski}
\usepackage[T1]{fontenc}
\usepackage{graphicx}
\usepackage{wrapfig}
\usepackage{txfonts}
\usepackage{amsmath}
\usepackage{float}
\usepackage{natbib}
\usepackage{graphicx}
\usepackage{txfonts}
\usepackage{amsmath}
\usepackage{float}
 \usepackage[table,xcdraw]{xcolor}
\usepackage[hidelinks,colorlinks=true,linkcolor=blue,citecolor=blue]{hyperref}
\DeclareUnicodeCharacter{2212}{-}
\begin{document} 

\UseRawInputEncoding

   \title{Finding strong gravitational lenses through self-attention}

   \subtitle{ Study based on the  Bologna Lens Challenge}

   \author{Hareesh Thuruthipilly
          \inst{1}
          \and
          Adam Zadrozny\inst{1}
          \and
          Agnieszka Pollo\inst{1,2}
          \and
          Marek Biesiada\inst{1,3}}

   \institute{National Centre for Nuclear Research, Warsaw, Poland\\
             \email{Hareesh.Thuruthipilly@ncbj.gov.pl},\\
             \email{Adam.Zadrozny@ncbj.gov.pl}, 
             \email{Agnieszka.Pollo@ncbj.gov.pl}
             \and
            Jagiellonian University, Krak\'{o}w, Poland 
            \and 
            Department of Astronomy, Beijing Normal University, Beijing 100875, China}

   \date{\today}

 
   \abstract
   {The upcoming large-scale surveys, such as the  Rubin Observatory Legacy Survey of Space and Time (LSST),  are expected to find approximately $10^5$ strong gravitational lenses by analysing data  many orders of magnitude larger than those in contemporary astronomical surveys. 
    In this case,  non-automated techniques will be highly challenging and time-consuming,  if they are possible at all.}
   {We propose a new automated architecture based on the principle of self-attention to find strong gravitational lenses. The advantages of self-attention-based encoder models over convolution neural networks (CNNs) are investigated, and ways to optimise the outcome of encoder models are analysed.}
   {We constructed and trained 21 self-attention-based encoder models and five CNNs to identify gravitational lenses from the Bologna Lens Challenge. Each model was trained separately using 18,000 simulated images, cross-validated using 2,000 images, and then applied to a test set with 100,000 images. We used four different metrics for evaluation: classification accuracy, the area under the receiver operating characteristic  (AUROC) curve, and   $TPR_0$   and  $TPR_{10}$ scores (two metrics of evaluation for the Bologna challenge). The performance of self-attention-based encoder models and CNNs participating in the challenge are compared.}
   {The encoder models performed better than the CNNs. They were able to surpass the CNN models that participated in the Bologna Lens Challenge by a high margin for the TPR$_0$ and TPR$_{10}$. In terms of the AUROC, the encoder models with 3 $\times 10^6$ parameters had equivalent scores to the top CNN model, which had around 23 $\times 10^6$ parameters.}
   {Self-attention-based  models have clear advantages compared to simpler CNNs. They perform  competitively  in comparison to the currently used residual neural networks. Self-attention-based models can identify  lensing candidates with a high confidence level and will be able to filter out potential candidates from real data. Moreover, introducing the encoder layers can also tackle the overfitting problem present in the CNNs by acting as effective filters. }


\titlerunning {SLs Through Self-Attention}
\authorrunning {Hareesh et al.}
   \maketitle
%

\section{Introduction}

Strong gravitational lensing is the phenomenon by which a distant galaxy or quasar produces multiple highly distorted images because of the gravitational field of the foreground galaxy or a nearby massive astronomical body. Finding and analysing these strong lenses (SLs) is important;  they  have diverse applications in cosmology and astrophysics, ranging from estimating the Universe's dark matter distribution to constraining the cosmological models \citep{Koopmans_2006,2009ApJ...691..531C,Collett_2014, Cao_2015, Bonvin_2016,2017MNRAS.465.4914B}. Consequently, the current and  upcoming surveys have given significant attention to detecting strong gravitational lensing systems. For a detailed review of the applications of strong lensing, we refer to \citet{1992ARA&A..30..311B} and \citet{2010ARA&A..48...87T}.

However, for all these analyses a large sample of SLs is required. Unfortunately, only a few hundred lensing systems have been detected and confirmed by the present astronomical surveys to date. One of the largest lens catalogues available now is from the Sloan Lens ACS Survey (SLACS), with only 130 observed lenses \citep{2008ApJ...682..964B}. With the upcoming era of advanced missions such as  Euclid \citep{scaramella2021euclid} and LSST \citep{Ivezi__2019,verma2019strong}, the number of observable SLs is expected to reach $10^5$, which should be identified from around $10^9$ astronomical objects. Similarly, the number of new SLs expected to be in the Square Kilometre Array (SKA) survey will  have similar orders of magnitude \citep{2015aska.confE..84M}. To analyse the enormous amount of data produced from the present and future large-scale surveys, various methods have been tried out, including  crowd science \citep{2016MNRAS.455.1171M} and semi-automated methods, for example  arc detectors \citep{2004A&A...416..391L,2007A&A...461..813C}. However, these methods have  only had minor success and were too time-consuming to be a practical proposition. Hence, the situation demands better and more effective alternative approaches to detect SLs in future large-scale surveys.

It is worth mentioning that the advancements in artificial intelligence (AI) have opened up a plethora of opportunities and have been widely applied in astronomy and astrophysics (e.g. galaxy classification by \citealt{2019PASP..131j8002P}, supernova classification by \citealt{2017ApJ...836...97C}, and lens modelling by \citealt{2019MNRAS.488..991P}).
A particular class of deep-learning models known as   convolutional neural networks (CNNs) has recently been shown to work exceptionally well in finding SLs. Hence, developing  deep-learning-based algorithms to detect SLs from  large-scale surveys is an actively investigated area now \citep{ Lanusse_2017, Schaefer_2018,2019MNRAS.487.5263D,2020MNRAS.496..381C}.
For instance, \citet{2017MNRAS.471..167J} applied CNNs to the data from the  Canada-France-Hawaii Telescope Legacy Survey (CFHTLS) to find SLs, and numerous other successful attempts of finding potential SL candidates from the Kilo Degree Survey (KiDS) have been reported \citep{2017MNRAS.472.1129P, Petrillo_2018,Petrillo_2019, He_2020,Li_2020}. Likewise, various groups have successfully used CNNs to identify strong lens galaxy-scale systems from large-scale surveys, such as the  Dark Energy Survey (DES) \citep{2019ApJS..243...17J, rojas2021strong}, the Dark Energy Spectroscopic Instrument Legacy Imaging Surveys  \citep{Huang_2020, Huang_2021}, and  Pan-STARRS \citep{Lens_CNN&A...644A.163C}, and from  comparatively small-scale surveys, such as VOICE \citep{2021arXiv210505602G}.

An exciting feature of the CNNs is that they can directly take the image as the input and learn the image features, making them one of the most popular and robust architectures currently being used. Generally, the learning capacity of a neural network increases with the number of layers in the network. The network can then learn the low-level features with the first layers and then learn more complex features with increasing depth \citep{russakovsky2015imagenet,simonyan2015deep}. However,  increasing the layers in the neural network will result in higher complexity, which in turn may lead to overfitting \citep{doi:10.1021/ci0342472}. In addition, the gradient of the cost function decreases exponentially. Eventually, it vanishes for very deep networks, commonly called the vanishing gradient effect \citep{Hochreiter:91, Hochreiter:01book}. These two problems meant that creating very deep Convolution Networks was a challenging task \citep{2015arXiv150500387S}. 

However, the recently introduced idea of residual learning tackles these problems by introducing skip connections between the input
and output of a few convolution layers \citep{he2015deep}. As a result, the CNN learns the difference between the inputs and outputs rather than their direct mapping. Due to the skip connections, the gradients can reach deeper layers, thus tackling the vanishing gradient effect.
Recently,   \citet{he2015delving} were able to build models as deep as 1000 layers while increasing classification accuracy for the ImageNet Large-Scale Visual Recognition Challenge 2015. However, the scientific community are constantly looking for alternative and simple solutions that can outperform the existing models with reasonable computational cost.

Recently, there was a breakthrough in natural language processing (NLP) by the introduction of a new self-attention-based architecture known as the transformers \citep{vaswani2017attention}. Since then, there have been attempts to adapt the idea of self-attention to build better image processing models \citep{ramachandran2019standalone,zhao2020exploring,tan2021explicitly}. The basic idea behind the transformer architecture is the attention mechanism, which has also found a wide variety of applications in machine learning \citep{zhang2019selfattention,fu2019dual}. In the case of NLP, self-attention correlates different positions of a single sequence in order to calculate a representation of the sequence.
 The idea of self-attention, as the name   suggests,  is to give relative importance to the input features based on the input features themselves, which helps the network to create a representation of the input with the relatively important features only. Recently, Facebook Inc. \citep{carion2020endtoend} and Google Brain \citep{dosovitskiy2020image} have been able to surpass the existing image recognition models with transformer-based architectures. 
To our best knowledge, the transformer-based models have not been employed in astrophysics yet.
In this paper we explore the possibilities of this new architecture in detecting strongly gravitationally lensed systems. 

We implemented various self-attention-based encoder models (transformer encoders) to find the gravitational lenses from the Bologna Lens Challenge. We also compared the performance of the encoder models with created CNNs, and the CNNs participated in the challenge. The main objective of our study was to explore how suitable the transformer encoders are  for finding strong lenses and how to optimise the performance of transformer encoders. From our analysis, we  found that the encoder models perform better than the CNN models compared. We were able to beat the top TPR$_0$ and TPR$_{10}$ score (two metrics of evaluation for the Bologna Challenge) by a significant margin and to reach the top AUROC reported during the challenge.

The paper is organised as follows. Section 2 briefly describes the data we used to train our models. Section 3 provides a brief overview of the methodology used in our study, including the model's architecture and information on how the models were trained. The results of our analysis are presented in section 4. A detailed discussion of our results with a brief review of the performance of encoder models compared to the CNN models that participated in the challenge is presented in section 5. Section 6 concludes our analysis by highlighting the advantages of the encoder models over CNN models.

\section{Data}
\begin{figure*}
\centering
\includegraphics[width=500pt,keepaspectratio]{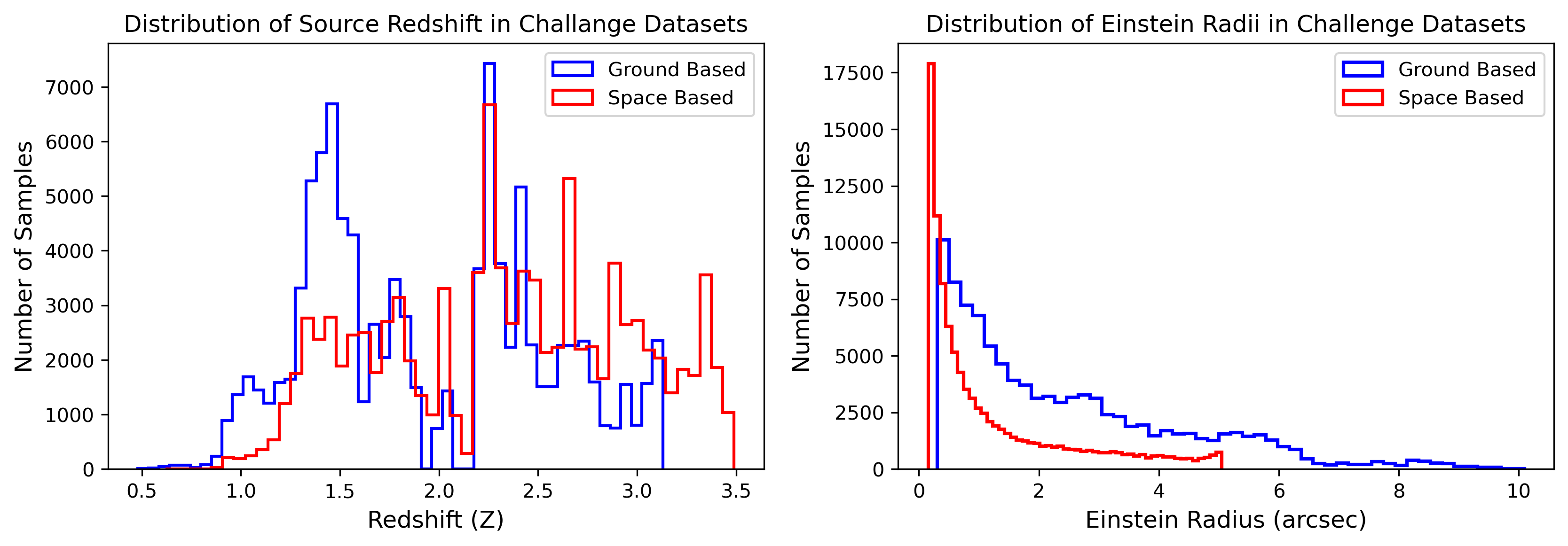}
\caption{Distributions of source redshifts and Einstein radii (in arcsec) of simulated gravitational lenses in the Bologna Lens Challenge.}
\label{fig:data_distribuyio}
\end{figure*}

The data used in this study is from the Bologna Strong Gravitational Lens Finding Challenge \citep{Metcalf_2019}. The challenge consisted of two different challenges that could be registered independently. The first challenge was designed to mimic the  datasets from surveys such as Euclid  consisting of single-band images. The second challenge was designed to resemble data from ground-based detectors with multiple bands. It was roughly modelled on the data from the Kilo-Degree Survey (KiDS) reported in \citet{2013ExA....35...25D}. However, the simulated images did not strictly mimic the surveys; they were only employed as references to set noise levels, pixel sizes, sensitivities, and other parameters. The distributions of source redshift and Einstein radii in the challenge datasets are shown in Fig. \ref{fig:data_distribuyio}. The challenge was opened on November 25, 2016, and closed on February 5, 2017. Surprisingly, automated methods such as CNN and SVM showed far better results than human inspection. During the challenge, these methods were able to classify the images with high confidence where a human would have doubt.  

The mock images for the challenge were created using Millennium simulation and GLAMER lensing code \citep{2009MNRAS.398.1150B,2014MNRAS.445.1942M}. Sources from the Hubble Ultra Deep Field (UDF) decomposed into shapelet functions were used to create the lensed background objects. There were 9,350 such sources with redshifts and separate shapelet coefficients in four bands.
The visible galaxies associated with the lens were simulated using an analytic model for the surface brightness of these galaxies. In particular, the S{\'e}rsic profile: $I(R) = I_0 \exp{-k R^{1/n_s} }$ was used. The parameters employed to simulate the galaxies were the total magnitude, the bulge-to-disc ratio, the disc scale height, and the bulge effective radius. The magnitude and bulge-to-disc ratio are a function of the passband. Each galaxy was provided with an inclination angle between $0^{\circ} $ and $80^{\circ} $ and random orientation. An elliptical S{\'e}rsic profile describes the bulge with an axis ratio randomly sampled between 0.5 and 1. The S{\' e}rsic index, $n_s$, is given by
\begin{equation}
    log(n_s) = 0.4 log \Big[max\Big(\frac{B}{T},0.03\Big)\Big]+0.1x,
\end{equation}
where x is a uniform random number between -1 and 1 and $B/T$ is the bulge-to-total flux ratio. The median redshift of sources in the space-based catalogue was $z_s = 2.35$ and in the ground-based catalogue it was $ z_s = 1.81$.

\subsection{Space-based}

The images for the space-based detector were simulated by \citet{Metcalf_2019} to roughly mimic the observations by the Euclid telescope in the visible channel. \citet{Metcalf_2019} set the pixel size as 0.1 arcsec, and applied a Gaussian point spread function (PSF)  with an FWHM of 0.18 arcsec to simulate the images. The reference band for background and foreground galaxies was the SDSS $i$, overlapping with the broader Euclid VIS band. The training set consisted of 20,000 images, and the challenge set consisted of 100,000 potential lens candidates.

\subsection{Ground-based}
\begin{figure}[h]
\centering
\includegraphics[width=250 pt,keepaspectratio]{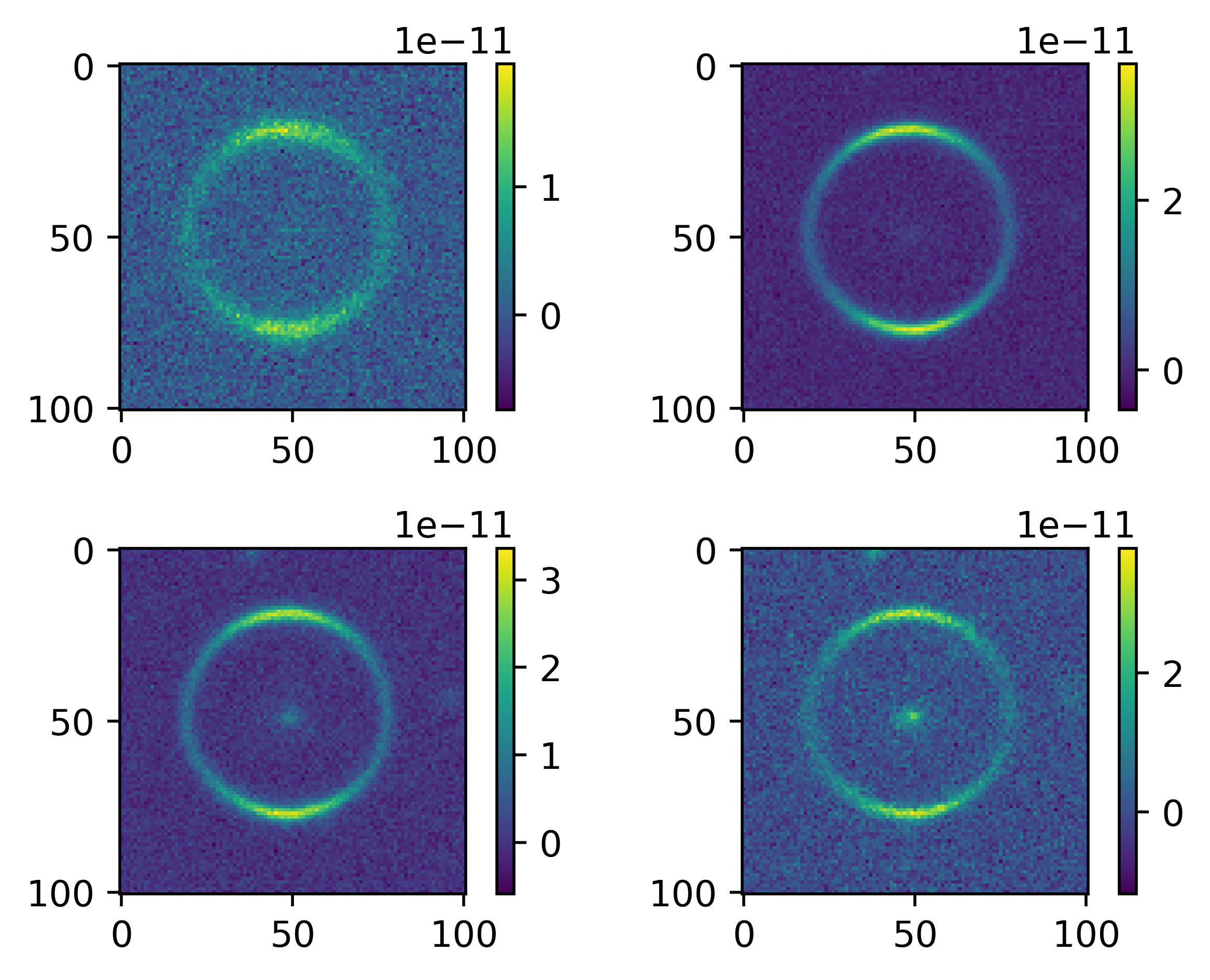}
\caption{Typical image of a mock simulated lens for the challenge. Bands are shown in the following order: $u$ (top left), $g$ (top right), $r$ (bottom left), and $i$ (bottom right).}
\label{fig:lenses}
\end{figure}

The ground-based images consisted of simulated images from four bands $(u, g, r,$ and $i)$, and the reference band was the $r$ band. In the challenge set, 85\% of the images were purely simulated. The other 15\%  were actual images chosen from a preliminary sample of bright galaxies directly from the KiDS survey. These real images were added to the challenge set for more realism. Some images had masked regions where removed stars, cosmic rays, and bad pixels were present. The noise for the mock images was simulated by adding normally distributed numbers with the variance given by the weight maps from the  KiDS survey. The example images of a mock simulated lens for the challenge are shown in Fig. \ref{fig:lenses}. For a detailed review of how the data was   created, we refer to \citet{Metcalf_2019}.

An exciting result reported from the challenge was that colour information was crucial for finding strong lenses. 
All the methods that participated in the challenge performed better on the data from the ground-based observatories, which had four photometric bands $(u, g, r, i)$, than on the data from the space-based detectors, which had a single band. 
Consequently, it was advised by \citet{Metcalf_2019} to add even low-resolution information from other instruments or telescopes to the higher resolution data in one band to improve the detection rates significantly. In other words, multiple bands make a significant difference, and future surveys will perform better if they have information provided in multiple bands. Hence, for our study, we chose the data from the ground-based observatories with  four photometric bands $(u, g, r, i)$ to study the attention-based models' ability to detect strong gravitational lenses. Since we are also interested in exploring the transformer architecture's optimisation and analysing the transformers' performance, a better data structure was preferred to compare transformer models.

\subsection{Data pre-processing}
The simulated datasets of the Ground-Based Bologna Strong Gravitational Lens Finding Challenge were provided in the FITS format and were available for download to the public \footnote{\hyperlink{http://metcalf1.difa.unibo.it/blf-portal/gg\_challenge.html}{\text{http://metcalf1.difa.unibo.it/blf-portal/gg\_challenge.html}}}. The challenge datasets contained $100,000$ potential strong lens candidates, and the training set contained 20000 images along with other information, such as the Einstein area in rad$^2$ and  number of pixels in the lensed image above $1 \times \sigma$. In this work we did not use additional information about the images. We only used whole images (101$\times$101 pixels) in all four photometric bands $(u, g, r, i)$ as an input to the model and information about the lens present or not as the desired output for training the models. During training the 20000 images were split into two parts. We used a dataset of 18,000 to train the network, which was used for validation. Before training the models, each image was re-scaled and rotated by $n \pi/2$, where $n \in (0,1,2,3),$ to enrich the dataset.

\section{Methodology}
\subsection{Convolution neural networks}
The concept of using CNNs to analyse image-like data was first proposed by  \citet{Lecun}. However, a breakthrough for image recognition by CNNs did not happen  until \citet{Imagenet} created an architecture that won the ImageNet Large-Scale Visual Recognition Challenge 2012. Since then, CNNs have been extensively employed in various research disciplines following the proposed architecture. A regular CNN can be thought of as consisting of two parts. The first part consists of convolution layers, and the second part of fully connected layers that resemble the usual artificial neural
networks (ANNs). The main advantage of using convolution layers is that they can learn the local spatial correlation in the data, so using multiple convolution layers will help us to detect the features in the data independent of their position \citep{Mallat_2016}.

During the training, the input image is convolved with a number of small kernels (or features maps, typically of dimension 3 $\times$ 3). These individual kernels are optimised during training. 
The final part of the CNN is the fully connected (FC) layers, which resemble the ANNs. They are used to consolidate the information contained in the feature maps to generate the output. However, a convolutional neural network is restricted by the size of its kernels to collect spatial information from the data. Hence it may lead to deviations due to the ignorance of global information.
Since the CNN models have high complexity and a large number of trainable parameters, they are usually prone to overfitting. In addition, the depth of the CNN models is limited by the vanishing gradient effect, where the gradient of the CNN layers vanishes as we go deeper.

\subsection{Self-attention}
The introduction of attention mechanisms in machine learning has the potential to revolutionise machine learning, and it has been found particularly useful in NLP. Depending on the task at hand, various attention mechanisms can be employed. Among them, self-attention is one of the highly used attention mechanisms for image analysis. For a review of various attention mechanisms, we refer to \citet{Yang_2020, NIU202148}.  During the application of self-attention, each point in the feature map generated by the convolution layer is considered a random variable, and the paring covariances are determined, so that the value of each prediction can be improved or minimised based on its similarity to other points in the feature map. In other words, the central idea of self-attention is to assign relative importance to the features of the input based on the input itself.

In general, the attention function can be defined mathematically as 
\begin{equation}\label{attention}
 \centering
 \text{Attention}(Q,K,V) = \text{softmax} \left( \frac{QK^T}{\sqrt{d_k}} \right) V,
\end{equation}
where $Q, K, V$ are vectors and $\sqrt{d_k}$ is the dimension of the vector key ($K$). The softmax function, by definition, is the normalised exponential function that takes an input vector of K real numbers and normalises it into a probability distribution consisting of K probabilities proportional to the exponentials of the input numbers.
As we compute the normalised dot product between the query ($Q$) and the key ($K$), we get a tensor ($QK^T$) that encodes the relative importance of the features in the key to the query \citep{vaswani2017attention}. For self-attention, the vectors ($Q$), ($V$), and ($K$) are identical. Hence multiplying the tensor ($QK^T$) by vector ($V$) results in a vector that encodes the relative importance of features inside the input vector.

\begin{figure}[h]
    \centering
    \includegraphics[width=250 pt,keepaspectratio]{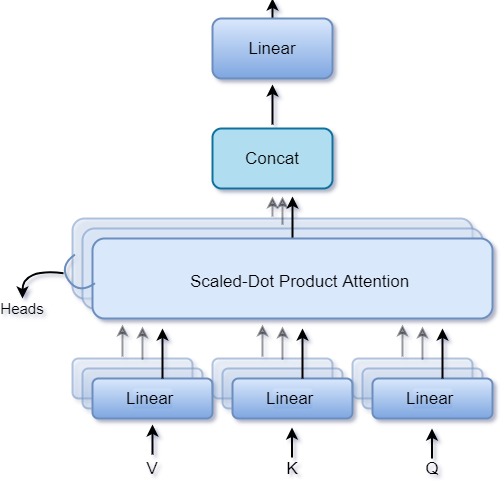} 
     \caption{Scheme of the multi-head attention layer.}
    \label{fig:Enocder}
\end{figure}

 A physical interpretation of self-attention applied to feature vectors can be thought of as filtering the input features based on the correlation in the input. The structure of a multi-head attention layer is given in Fig. \ref{fig:Enocder}. It is possible to provide the self-attention with more power by creating several layers and dividing the input vector into smaller parts (H, number of heads). Each attention layer is called a head, which applies self-attention to one part of the divided input. 

 \subsection{Positional encoding}
 If we pass the input directly to the attention layers,  the input order or the positional information is lost as transformer models are permutation invariant. So to preserve the information regarding the order of features, we use positional encoding, and the lack of positional encoding will lower the performance of a transformer model. Following the work of \citet{vaswani2017attention},  we used fixed positional encoding defined by the function 
 \begin{gather}
     PE_{(pos,2i)} = \rm sin \Big(pos/12800^\frac{2i}{d_{model}}\Big),\\
     PE_{(pos,2i+1)} = \rm cos \Big(pos/12800^\frac{2i}{d_{model}}\Big),
 \end{gather}
 where $pos$ is the position, $i$ is the dimension, and $d_{model}$ is the dimension of the input feature vector. Each dimension of the positional encoding corresponds to a sinusoid function. For a detailed description of positional encoding and its importance  we refer to \citet{vaswani2017attention,liutkus2021relative,su2021roformer,chen2021demystifying}.

\subsection{Transformer encoder}
\begin{figure}[h]
\centering
\includegraphics[width=250 pt,keepaspectratio]{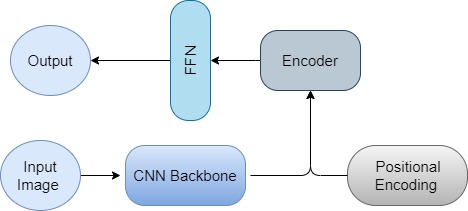}
\caption{Scheme of the  architecture of the transformer encoder}
\label{fig:Transformer}
\end{figure}

The Transformer models we constructed to detect SLs were inspired by the DEtection TRansformer (DETR) created by Facebook \citep{carion2020endtoend}. As shown in Fig. \ref{fig:Transformer}, the transformer encoder has a very simple architecture and contains three main parts, which are desctibed below.

    The first component of the architecture is a simple CNN used to extract the features from the image. The output from the CNN backbone is   a vector with dimensions H$\times$W$\times$D, where D is the number of filters in the last convolution layer. The encoder demands a sequence as input, hence we have to reshape the output of the CNN to a D$\times$HW feature map.
    As mentioned above, the transformer architecture is permutation-invariant, hence for the second component we add the output of the CNN backbone with fixed positional encoding before processing it to the transformer encoder layer. After the CNN backbone, we process the self-attention-based encoder layers,   and filter the relevant features extracted by the CNN. The encoder layer has a standard architecture and consists of a multi-head self-attention module and a feed-forward network (FFN).  
    The third part of the model is a  FFN that is  similar to the regular CNNs and that learns the features filtered by the encoder layers. The model's output is a single neuron with a sigmoid activation function that predicts the probability that the input image is a lens.

 We created 21 encoder models with different structures to study how the hyperparameters in the encoder will affect the model's performance. We used the exponential linear unit (ELU) function as the activation function for all the layers in these models. We initialise the weights of our model with the  Xavier uniform initialiser, and all layers are trained from scratch by the ADAM optimiser with the default exponential decay rates \citep{Glorot2010UnderstandingTD,kingma2017adam}. 
 
 \subsection{Lens detector}\label{section:Lens_Detector}

Among the created encoder models, the best performance was given by the encoder model that  uses a CNN backbone similar to the LASTRO-EPFL model, from the Bologna Lens Challenge \citep{Schaefer_2018}. Here we present the two  best  architectures: Lens Detector 15 and Lens Detector 16, which outperformed all the other models during our analysis. 
\begin{figure*}\centering
\includegraphics[width=450 pt]{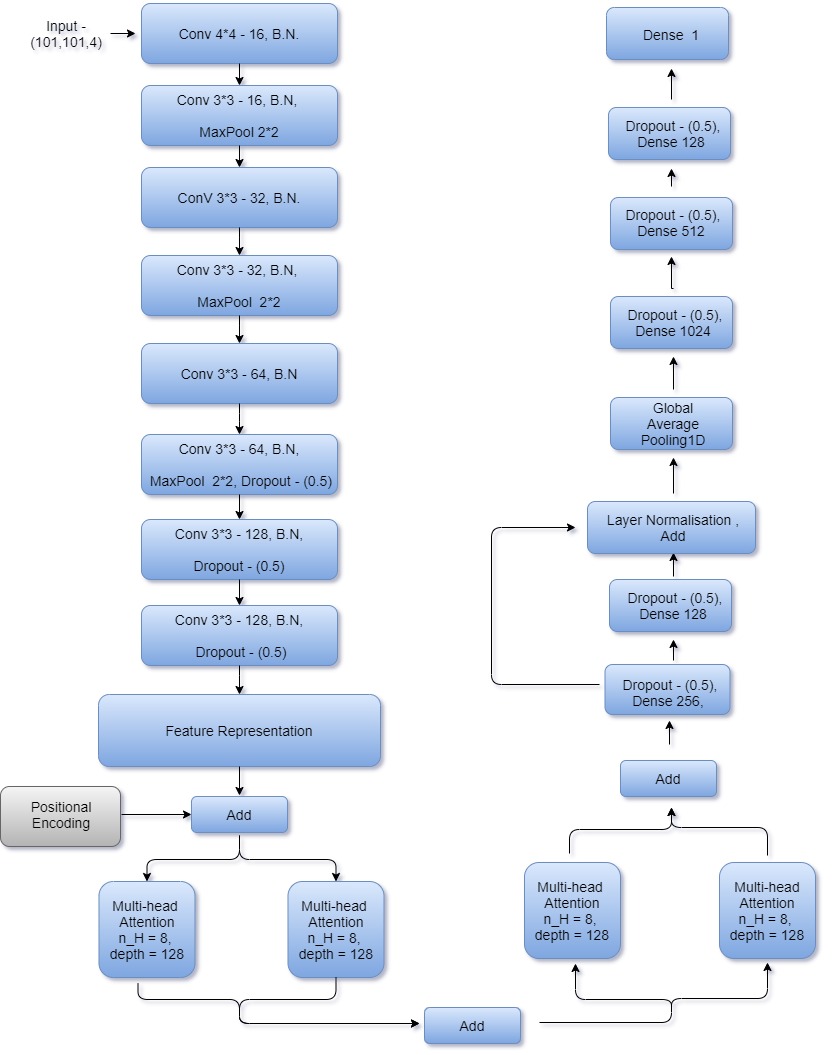}
\caption{Scheme of the  architecture of  Lens Detector 15}
\label{fig:Lens_Detector}
\end{figure*}
  Lens Detector 15  was first trained for 300 epochs with an initial learning rate of $\alpha = 10^{-4}$ and again trained for another 100 epochs starting with a learning rate of $\alpha = 10^{-5}$. This version of the lens detector gave high scores in all three evaluation metrics for the challenge. The architecture of Lens Detector 15 is given in Fig. \ref{fig:Lens_Detector}. In the spirit of reproducible research, our code for Lens Detector 15 is publicly available\footnote{\hyperlink{https://github.com/hareesht23/Lens-Detector}{https://github.com/hareesht23/Lens-Detector}}.

\begin{figure}[h]\centering
\includegraphics[width=250 pt]{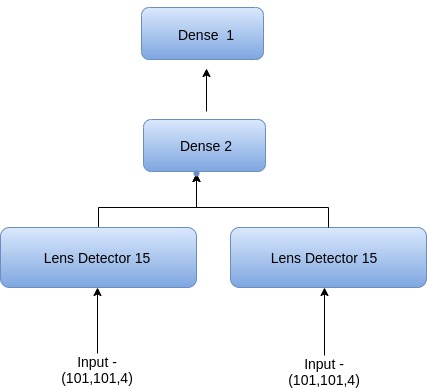}
\caption{Scheme of the  architecture of  Lens Detector 16}
\label{fig:Lens_Detector 16}
\end{figure}

Lens Detector 16 was created by stacking two Lens Detector 15 models in parallel and combining their outputs through an additional dense layer connected to a single neuron to give the output. The architecture of   Lens Detector 16  is shown in Fig. \ref{fig:Lens_Detector 16}. 
  Lens Detector 16 was first trained for 100 epochs with an initial learning rate of $\alpha = 10^{-4}$ and again trained for another 100 epochs starting with a learning rate of $\alpha = 5\times 10^{-5}$. Furthermore, the model was trained for 50 epochs with $\alpha = 10^{-5}$ and after that with $ \alpha = 5\times 10^{-6}$ for another 200 epochs.  

We created the  Space Lens Detector model to identify strong lenses from the space-based dataset. The Space Lens Detector has a similar structure to Lens Detector 15. The only difference is the use of four heads in the encoder layers. The model was trained with an initial learning rate of $\alpha = 10^{-4}$ using the ten-fold validation and iterating for 20 epochs in each fold. 

\subsection{Metrics for evaluation}
We first start with a brief overview of the performance metrics we
used to quantify the performance of the lens classification. We used classification accuracy as the metric to compare the created transformer models. The classification accuracy is calculated as 
\begin{equation}
    \text{Accuracy} = \frac{TP+TN}{TP+FP+TN+FN},
\end{equation}
where TP is the number of true positives, TN is the number of true negatives, FP is the number of false positives, and FN is the number of false negatives. 
Apart from the classification accuracy, another popular figure of merit for a classifier is the area under the receiver operating characteristic (AUROC) curve \citep{Metcalf_2019}. The receiver operating characteristic curve is created by plotting the true positive rate (TPR) against the false positive rate (FPR) as a function of the threshold. The TPR is the ratio of detected lenses to the total number of lenses:\begin{equation}
    TPR = \frac{TP}{TP + FN}.
\end{equation}
The  TPR measures how well the classifier detects lenses from the whole population of objects.
The FPR can be understood as a contamination rate in the classification and defined as the fraction of non-lens images incorrectly identified as lenses: 
\begin{equation}
    FPR = \frac{FP} {FP + TN}.
\end{equation} 
The AUROC assesses the overall ability of a classifier to distinguish between classes. A perfect classifier will have AUROC = 1.0 with TPR = 1.0 and FPR = 0.0 for any threshold, whereas a random classifier will have AUROC = 0.5 with TPR = FPR for any threshold.   For the Bologna Lens Challenge the participants were instructed to optimise AUROC rather than the accuracy. In addition, two more figures of merit were also considered for the competition, which are TPR$_0$ and TPR$_{10}$. 
The TPR$_0$ is defined as the highest TPR reached, as a function of the p threshold, before a single false positive occurs in the test set of 100,000 cases. This is the point where the ROC meets the FPR = 0 axis. If the classifier assigns a high probability for a non-lensed image to be a lens, even for one case, the TPR$_0$ will go low. This means that the TPR$_0$ parameter measures the confidence in the purity of the samples identified by a model. Similarly, the  TPR$_{10}$  is defined as the TPR at the point where fewer than ten false positives are made, which is also a measure of confidence in the true samples mined out by a model with a slight impurity. A high TPR$_0$ and TPR$_{10}$ indicate that the classifier can clearly  distinguish between lensed and non-lensed images. 

\section{Results}
We created 5 convolution models to use as the backbones for the encoder models and 21 encoder models to study how the hyperparameters of the encoder layer affect the performance. Since each architecture was implemented as a regression model, a probability of 0.5 was set as the threshold for classifying an image as a lens or not. Thus, input images with a prediction value less than 0.5 were classified as non-lensed images labelled zero and vice versa. Table \ref{table:1} describes the architecture and total accuracy,  AUROC, TPR$_0$, and TPR$_{10}$ of all created models.

\begin{table*}
\centering 
\addtolength{\tabcolsep}{4pt}
\resizebox{\width}{!}{
\begin{tabular}{c c c c c c c}\hline\hline
\textbf{Model   name} & \textbf{Model structure} &\textbf{Accuracy}& \textbf{AUROC} & \textbf{TPR$_0$} & \textbf{TPR$_{10}$} \\\hline
CNN 1    & 5 CNN Layers & 88.21 & 0.951 & 0.000 & 0.07 \\
CNN 2    & 4 CNN Layers  & 86.74 & 0.915 & 0.000 & 0.4  \\
CNN 3    & 8 CNN Layers     & 88.51 & 0.968 & 0.033 & 0.37  \\
CNN 4    & 3 CNN Layers    & 88.49& 0.956 & 0.000 & 0.68   \\
CNN 5    & 25 CNN Layers    & 91.26 & 0.974 & 0.004 & 0.004   \\
Lens   Detector 1       & CNN 1+1 H$_{16}$+1(E)     & 89.57 & 0.961 & 0.000 & 0.643  \\
Lens   Detector 2       & CNN 2 + 1 H$_{16}$ + 1(E)  &88.13 & 0.950 & 0.001 & 0.001  \\
Lens   Detector 3       & CNN 2 + 2 H$_{16}$ + 1(E) & 88.00 & 0.962 & 0.018 & 0.018  \\
Lens   Detector 4       & CNN 2 + 2 H$_{32}$ + 1(E)               & 88.12& 0.952 & 0.121 & 0.124 \\
Lens   Detector 5       & CNN 2 + 4 H$_{64}$ + 2 (E)            & 88.46 & 0.955 & 0.125 & 0.133 \\
Lens   Detector 6       & CNN 2 + 4 H$_{128}$ + 4(E)           & 89.51& 0.957 & 0.003 & 0.004  \\
Lens   Detector 7       & CNN 3  + 8 H$_{128}$ + 2(E)  & 91.45 & 0.968 & 0.000 & 0.410   \\
Lens   Detector 8       & CNN 4 + 2 H$_{384}$ + 2 (E)     & 89.43 & 0.954 & 0.000  & 0.758  \\
Lens   Detector 9       & 3 CNN Layers + 2 H$_{384}$ + 2 (E)  & 89.61 & 0.959 & 0.000  & \textbf{0.789} \\
Lens   Detector 10      & 5 CNN Layers + 8 H$_{128}$ + 2 (E) & 90.58& 0.970 & 0.180 & 0.23 \\
Lens   Detector 11      & 5 CNN Layers + 8 H$_{128}$ + 4 (E)  & 90.45 & 0.966 & 0.219 & 0.34  \\
Lens   Detector 12      & 8 CNN Layers + 8 H$_{128}$ + 4 (E)    & 89.82  & 0.960 & 0.040 & 0.680 \\
Lens   Detector 13      & 8 CNN Layers + 8 H$_{128}$ + 4 (E) & 91.94& 0.975 & 0.175 & 0.525  \\
Lens   Detector 14      & 8 CNN Layers + 8 H$_{128}$ + 4 (E)  & 91.95 & 0.975 & 0.002 & 0.539  \\ 
Lens   Detector 15      & 8 CNN Layers + 8 H$_{128}$ + 4 (E)  & \textbf{92.99} & 0.978  & 0.140  & 0.48  \\
Lens   Detector 16      & 16 CNN Layers + 8 H$_{128}$ + 8 (E)    & 90.97  & 0.962 & \textbf{0.225} & 0.24  \\
Lens   Detector 17      & 16 CNN Layers + 8 H$_{128}$ + 8 (E) & 92.19  & 0.973 & 0.00  & 0.717   \\
Lens   Detector 18      & 16 CNN Layers + 8 H$_{128}$ + 8 (E)  & 92.09  & 0.976& 0.113  & 0.590  \\
Lens   Detector 19      & 16 CNN   Layers + 16 H$_{128}$ + 8 (E) & 90.03   & 0.961 & 0.114 & 0.115 \\
Lens   Detector 20      & 25 CNN Layers + 8 H$_{128}$ + 4 (E) & 91.26 & 0.973 & 0.212 & 0.223  \\
Lens   Detector 21      & 8 CNN Layers + 8 H$_{128}$ + 4 (E) & 92.79 & \textbf{0.98}& 0.00  & 0.64   \\
 \hline
\end{tabular}}
\caption{Table comprising the architecture, accuracy, AUROC, TPR$_0$, and TPR$_{10}$  of all the models in  chronological order of creation. The encoder models are named `Lens Detector' followed by a number. The model structure describes if the model uses transfer learning in the CNN backbone or not. Generally, the term `$
X H_{Y}$' in the model structure means there are x heads with dimension y in one encoder layer. Similarly, the  term `$Z(E)$' denotes that there are Z   encoders in the structure.}
\label{table:1}
\end{table*}

\begin{figure}[h]
    \centering
    \includegraphics[width=250 pt,keepaspectratio]{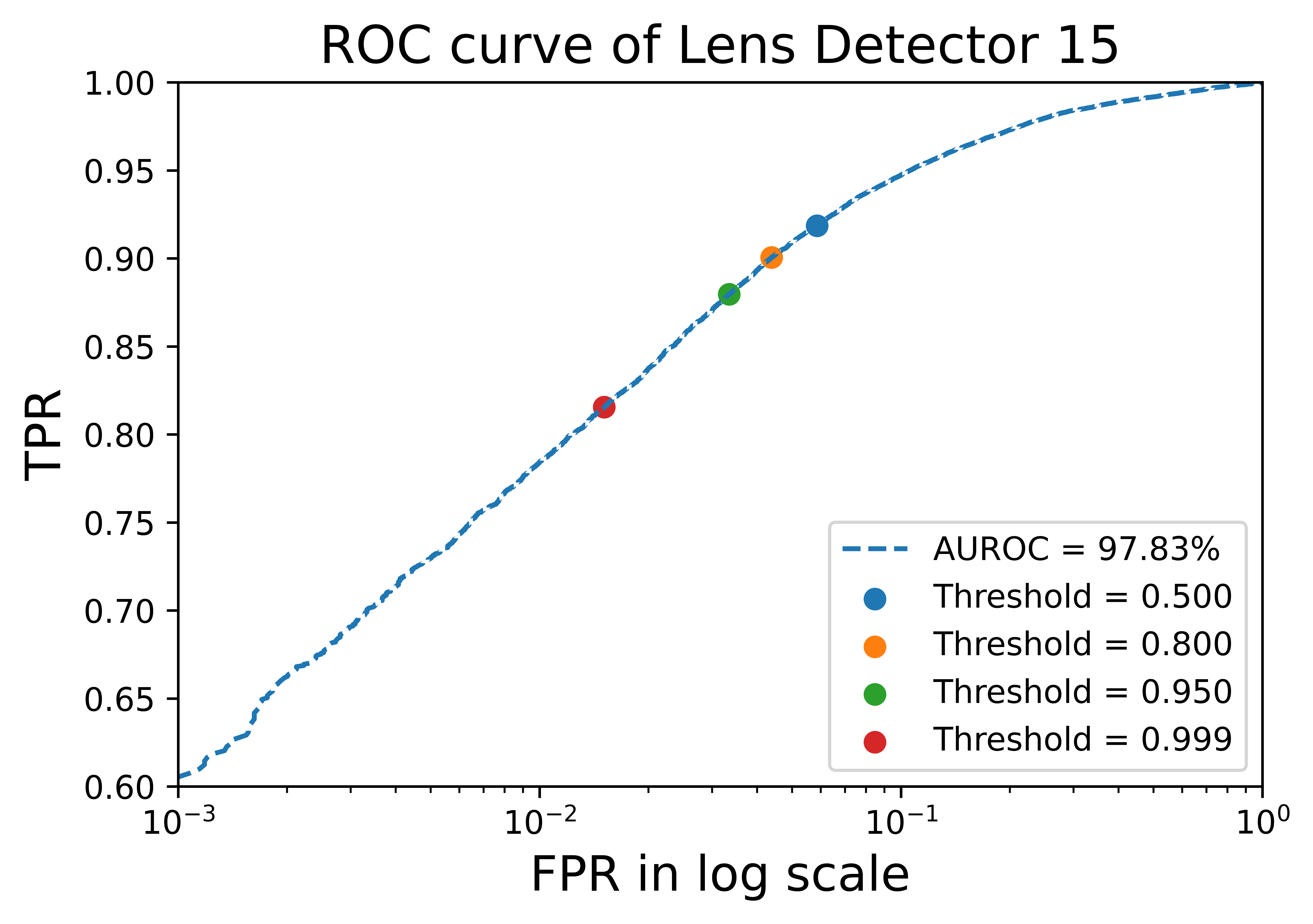}
    \caption{Receiver operating characteristic (ROC) curve of  Lens Detector 15.}
    \label{fig:ROC}
\end{figure}

Among the created encoder models the highest accuracy was achieved by Lens Detector 15 and the highest AUROC, TPR$_0$, and TPR$_{10}$ were achieved by   Lens Detector 21,  Lens Detector 16, and Lens Detector 9, respectively. From the presented models here, we would like to highlight     Lens Detector 15 as the best model since it performs well in all categories and has the highest classification accuracy. The receiver operator characteristic (ROC) curve of  Lens Detector 15 is shown in Fig. \ref{fig:ROC}. Similarly,  Lens Detector 13, Lens Detector 18, and Lens Detector 20  can also be considered  highly performing classifiers. All of these models scored an AUROC equivalent to the second-best model that participated in the challenge and a better TPR$_0$ and TPR$_{10}$ compared to all other models that participated in the challenge. The ROC curves of all the encoder models are presented in Fig. \ref{fig:my_label} in Appendix A.

\begin{figure*}
    \centering
    \includegraphics[width=500 pt]{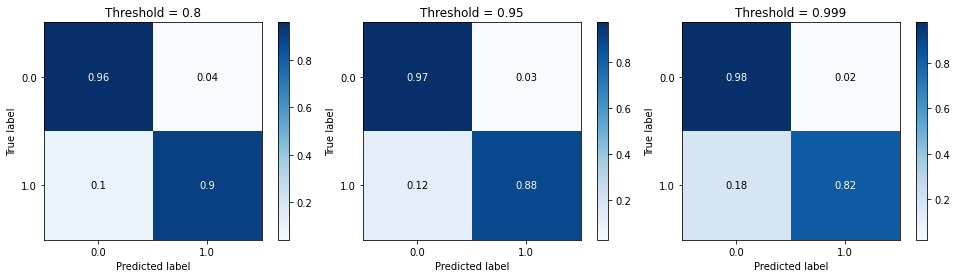}
    \caption{Confusion matrix of Lens Detector 15 plotted for various thresholds. Class 0 represents the non-lensed images, and Class 1 represents the lensed images. The lower right square in each confusion matrix represents the true positives  for which  Lens Detector 15 identified strong lenses correctly. The upper left square in each confusion matrix represents the true negatives for which  Lens Detector 15 identified non-strong lenses correctly. The lower left square in each confusion matrix represents the false negatives or the missed true lenses by Lens Detector 15. The upper right square in each confusion matrix represents the false positives or the non-lenses identified by  Lens Detector 15 as strong lenses. }
    \label{fig:CM15}
\end{figure*}

Even though we  chose 0.5 as the threshold for identifying a candidate as a strong lens, in reality such a threshold is not  practical since the number of lenses to visually inspect after the run of the network could be unrealistically high. In order to validate the performance of  Lens Detector 15, we   plotted the confusion matrix of the lens detector with varying thresholds (see Fig. \ref{fig:CM15}). For the ground-based data, we were able to mine out more than 80\% of the true lenses with a threshold as high as 0.999.

\begin{figure}[h]
    \centering
    \includegraphics[width=250 pt]{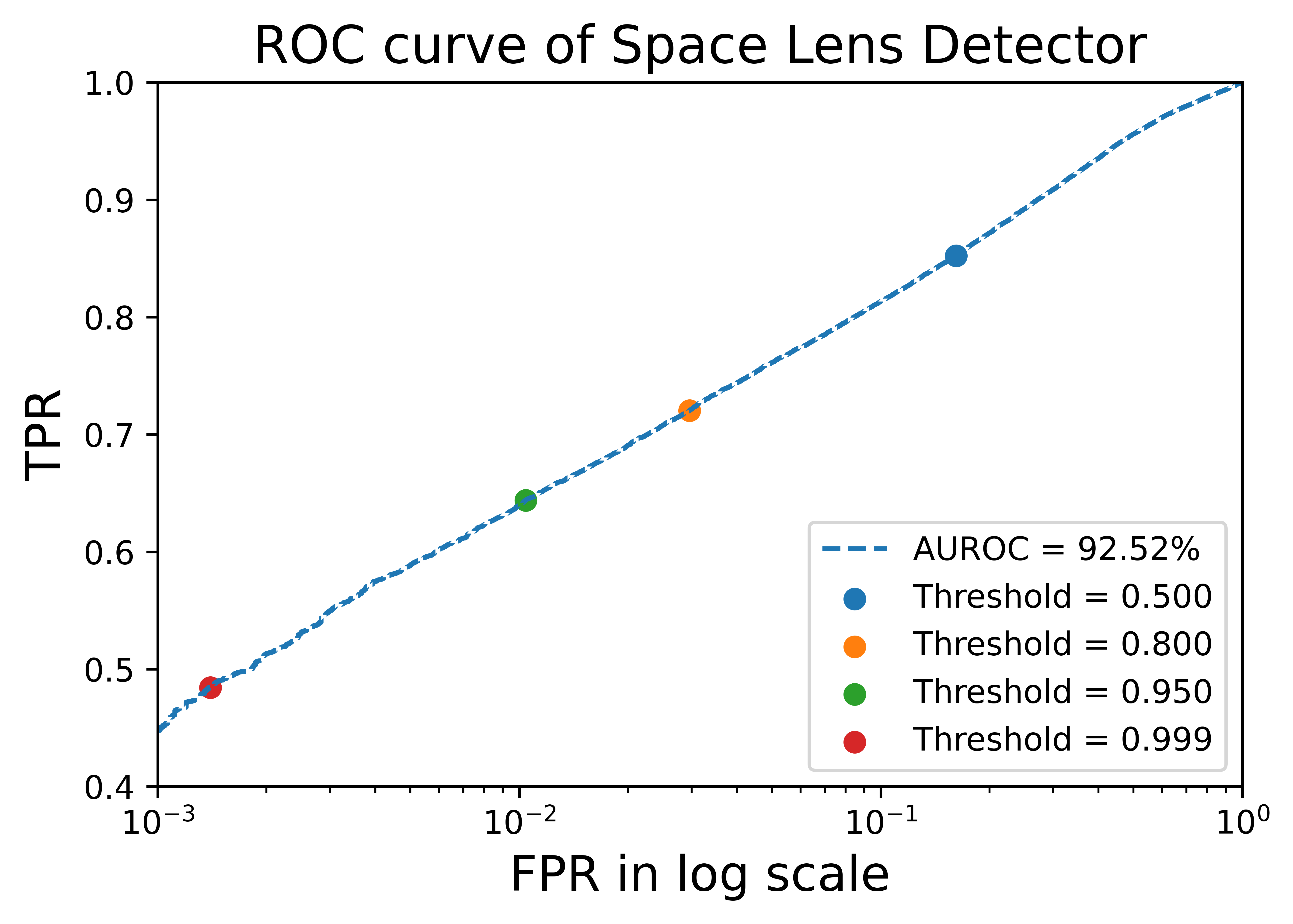}
    \caption{ROC curve of the Space Lens Detector.} 
    \label{fig:S_ROC}
\end{figure}

To identify strong lenses from the space-based dataset, we created the Space Lens Detector, which scored an AUROC = 0.925. The corresponding AUROC is greater than the second-best AUROC reported in the Bologna Lens Challenge and a little below the top AUROC (0.93) of the Bologna Lens Challenge. The Space Lens detector scored a TPR$_0$ of 0.039 and TPR$_{10}$ of 0.166, which is comparable to the performance of the other machine learning techniques that participated in the challenge. Since it is    clear that the attention-based encoder models can identify the SLs from single-band images, we did not try to improve the scores further. The ROC curve of the Space Lens Detector is shown in Fig. \ref{fig:S_ROC}. 

Similarly to the models that participated in the challenge, encoder models also performed better on the data from the ground-based observatories, which had four photometric bands $(u, g, r, i)$, compared to the data from the space-based detectors, which had a single band. Our results support the argument presented in \citet{Metcalf_2019} that multiple bands make a significant difference and improve the detection.
\section{Discussion}

\subsection{Transformers and models from the  Bologna Lens Challenge}
The Bologna Lens Challenge was intended to improve the efficiency and biases of tools used to find strong gravitational lenses on galactic scales. It was clear from the challenge that automated methods such as CNNs and SVM have a clear advantage compared to conventional methods. The performance of all these models was evaluated using AUROC, TPR$_0$, and TPR$_{10}$ scored on the challenge set. An SVM model named Manchester SVM won the competition in the  TPR$_0$ category with a score of 0.22. \citep{Metcalf_2019, Hartley_2017}. The model named CMU Deep Lens received  an AUROC of 0.981 and a TPR$_{10}$ score of 0.45,  the highest in their respective categories, and thus won the competition \citep{Metcalf_2019, Lanusse_2017}. Another variant of the model, named CMU Deep Lens, also received an AUROC of 0.98 during the challenge. These models were 46 layers deep ResNets with around 23 $\times 10^6$ parameters \citep{Lanusse_2017}. Another model worth mentioning from the challenge is  LASTRO   EPFL, an eight-layer CNN that won the competition for the space-based dataset in the AUROC category. For a detailed look at the models that participated in the challenge we refer to \citet{Hartley_2017} for the SVM, \citet{Lanusse_2017} for CMU-DeepLens, and \citet{Schaefer_2018} for LASTRO   EPFL.

We would like to compare the performance of these models to the performance of the encoder models to exhibit the advantages of encoder models over the  CNNs and SVM models. As mentioned earlier, we focused on the data from ground-based observatories. Here we are only comparing the performance of the created encoder models and that of the models that participated in the challenge only for the ground-based observatories data. The values reported in \citet{Metcalf_2019} are used here for the comparison. 

During the challenge, the TPR$_0$ was used to strongly penalise the classifiers with discrete ranking because their highest classification level was not conservative enough to eliminate all false positives, and they were likely to get TPR$_0 = 0$. For the other models that participated in the challenge, maximising the TPR$_0$ was a tough challenge, also for encoder models. However, the encoder models performed very well compared to the CNN models that participated in the challenge. The results of TPR$_0$ for the top three encoder models and the top three models that participated in the challenge are listed in Table \ref{table:Trp0}. We would like to note two models; Lens Detector 16 achieved a TPR$_0$ of \textbf{0.225}  and Lens Detector 11 reached a TPR$_0$ of 0.219, which are very high compared to the CNNs that participated in the challenge.

\begin{table}[H]
\centering\addtolength{\tabcolsep}{-0.5pt}
\resizebox{\width}{!}{
\begin{tabular}{ccccc} \hline\hline
\textbf{Name} & \textbf{AUROC} & \textbf{TPR$_0$} & \textbf{TPR$_{10}$} & \textbf{Model type} \\ \hline
Lens   Detector 16         & 0.962 & 0.225 & 0.24 & Transformer \\ 
Manchester   SVM            & 0.93   & 0.220   & 0.35 & SVM/Gabor   \\
Lens   Detector 11          & 0.966 & 0.219 & 0.34 & Transformer \\
Lens   Detector 15         & 0.978  & 0.140  & 0.48 & Transformer \\
\shortstack{CMU-DeepLens \\Resnet-ground3} & 0.98   & 0.09   & 0.45 & CNN         \\
LASTRO   EPFL               & 0.97   & 0.07   & 0.11 & CNN         \\\hline
\end{tabular}}
\caption{Comparison of encoder models and models that participated in the Bologna Lens Challenge, listed in  decreasing order of $TPR_0$.}\label{table:Trp0}
\end{table}

 \begin{figure*}
    \centering
\includegraphics[width=500 pt,keepaspectratio]{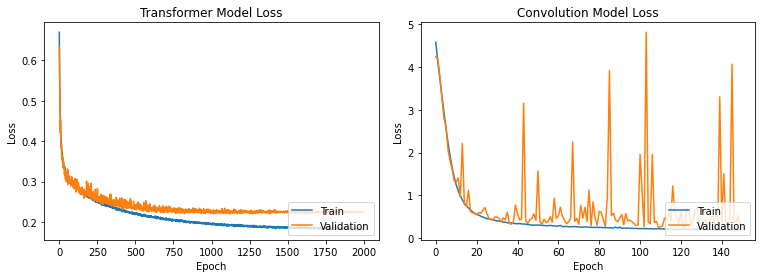} 
     \caption{Variation of loss function with epochs for Lens Detector 13 and CNN 3, respectively. Lens Detector 13 uses CNN 3 as its CNN backbone.}
    \label{fig:Loss transformer} 
\end{figure*}

The next parameter used to evaluate the models in the challenge was TPR$_{10}$ for which the encoder models showed a high range of supremacy over all other models that participated in the challenge. Particularly,  Lens Detector 9 achieved   TPR$_{10}$ = 0.79. The results of TPR$_{10}$ for the top three encoder models and the top three models that participated in the challenge are listed in Table \ref{table:trp10}. Three of our models were able to score a TPR$_{10}$ above 0.70, which is very high compared to the top TPR$_{10}$ reported during the challenge.  Table \ref{table:1} clearly shows that most encoder models achieved a higher score in this category compared to the other models that participated in the challenge. 

\begin{table}[H]
\centering\addtolength{\tabcolsep}{-0.5pt}
\resizebox{\width}{!}{
\begin{tabular}{ccccc} \hline\hline
\textbf{Name} & \textbf{AUROC} & \textbf{TPR$_0$} & \textbf{TPR$_{10}$} & \textbf{Model type} \\ \hline
Lens   Detector 9          & 0.959 & 0.00  & 0.789 & Transformer \\ 
Lens   Detector 8          & 0.954 & 0.00  & 0.758 & Transformer \\ 
Lens   Detector 17         & 0.973 & 0.00  & 0.717 & Transformer \\ 
CMU-DeepLens & 0.98   & 0.09   & 0.45 & CNN         \\
Resnet-ground3 & & & & \\
Manchester   SVM            & 0.93   & 0.220   & 0.35 & SVM/Gabor   \\
LASTRO   EPFL               & 0.97   & 0.07   & 0.11 & CNN         \\ \hline
\end{tabular}}
\caption{Comparison of encoder models and models that participated in the Bologna Lens Challenge, listed in   decreasing order of $TPR_{10}$.}\label{table:trp10}
\end{table}

 Now looking at the third parameter of merit used in the Bologna Lens Challenge, which is the AUROC, we can see that  Lens Detector 21 was able to reach the highest reported AUROC in the Bologna Lens Challenge \citep{Metcalf_2019}. The top three encoder models and the top three models that participated in the challenge that scored the highest AUROC are listed in Table \ref{table:auroc}.  
However, the CMU-DeepLens, was a 46 layer deep ResNet with around 23 $\times 10^6$ parameters \citep{Lanusse_2017}, whereas  Lens Detector 21 had only 3 $\times 10^6$ parameters and achieved an AUROC of 0.9809, which is very close to the performance of CMU Deep Lens (AUROC = 0.9814).

\begin{table}[H]
\centering\addtolength{\tabcolsep}{-0.5pt}
\resizebox{\width}{!}{
\begin{tabular}{ccccc} \hline\hline
\textbf{Name} & \textbf{AUROC} & \textbf{TPR$_0$} & \textbf{TPR$_{10}$} & \textbf{Model type} \\ \hline
CMU-DeepLens  & 0.981   & 0.09   & 0.45 & CNN         \\
Resnet-ground3 & & & & \\
Lens   Detector 21          & 0.981 & 0.00  & 0.64 & Transformer \\
CMU-DeepLens  & 0.980   & 0.02   & 0.10 & CNN         \\
Resnet-Voting & & & & \\
Lens   Detector 15          & 0.978 & 0.140  & 0.48 & Transformer \\
Lens   Detector 18          & 0.976& 0.113  & 0.59 & Transformer \\
LASTRO   EPFL               & 0.97   & 0.07   & 0.11 & CNN         \\\hline
\end{tabular}}
\caption{Comparison of encoder models and models that participated in the Bologna Lens Challenge, listed in   decreasing order of AUROC.}\label{table:auroc}
\end{table}

\subsection{Insights into transformers}
An initial glance at the results in Table \ref{table:1} shows that encoder models perform better than CNN models. However, the encoder models depend on the CNN backbone to extract the features, and as a result the performance of the encoder models depends upon the CNN backbone. A detailed look at the results indicates that the encoder model is only good as its CNN backbone. However, the encoder model always performs better than its CNN backbone. For example, the lowest accuracy achieved among the encoder models was for Lens Detector 3, which has better performance than CNN 2. A similar trend can be observed for other encoder models, which use trained CNNs as their backbone and perform better than the trained CNNs. These observations show that the encoder models can achieve  accuracy that is better by a small percentage than a CNN with the same number of convolution layers. 

However, an interesting question that should be addressed is what happens if we use deeper CNN backbones. We  compare the performance of a deeper CNN and an encoder model with a deeper CNN backbone. Model CNN 5 with 25 convolution layers and Lens Detector 20 with the same number of convolution layers give  \textbf{same} AUROC (0.97) and the same accuracy (0.91). We can see that the two models perform equally well. Using self-attention in deeper CNNs did not significantly improve the AUROC or accuracy. However, the TPR$_0$ score and TPR$_{10}$ score of   Lens Detector 20 ($0.212$ and $0.223$, respectively) is higher compared to  CNN 5 ($0.004$ and $0.004$, respectively). One probable reason why self-attention does not improve the accuracy and AUROC  of deeper CNNs is that the CNN backbone will learn more about the image's micro-scale features in deeper layers. Hence, the model will miss the long-range correlations of the original image found by the encoder layer.

In this section we speak of  the number of hyper-parameters in the encoder layer. On analysing the results from the encoder models, we can see that increasing the number of heads and the depth of the encoder increases the model's performance. However, increasing the number of heads and the depth of the encoder also increases the number of trainable parameters in the model. During the training period, it was found that increasing the number of parameters in the encoder layer helps the model to learn faster. This points to an exciting aspect of the encoder models. The encoder's performance is proportional to the number of trainable parameters in the encoder layer or specifically in the multi-head attention layer. The higher the number of trainable parameters, the better the learning curve and performance. However, the performance saturates beyond a limit for a given CNN backbone. 

Another interesting observation was the difference in the number of trainable parameters in the CNNs and self-attention-based encoder models. For example, an eight-layer CNN will have 4$ \times 10^6$ parameters, whereas the encoder with eight CNN layers and four encoder layers with eight heads each will have 3$\times 10^6$ parameters. In CNNs, most of the parameters come from the connections between the flattened output of the last convolution layer to the following dense layer. The weights in these layers help the CNNs to learn the features of the image. However, for a transformer network most of the trainable parameters are in the attention layers, which are only trying to learn the long-range correlations in the data and effectively act as a filter. This is one of the reasons why transformer networks can prevent overfitting. However, transformer models did not show any advantages over CNNs for the time taken to test and train the models. 
 
We also tried different learning rates from different ranges to find the optimal learning rate. We found   that increasing the learning rate above 0.001 considerably reduced the performance of the lens detectors. With an initial learning rate of 0.01, the models were not able to learn the features of the lenses from the training set. The optimal learning rate for our model was found to be  between $\alpha = 5 \times 10^{-5}$ and $\alpha = 2\times 10^{-4}$;  we chose $\alpha = 10^{-4}$ as the initial learning rate for our models.

Another striking feature worth pointing out is that, unlike convolution layers increasing the number of parameters in the encoder layers has a very slight effect on  overfitting the model.
Since the self-attention layers act as the filters for features extracted by CNN, an increase in the number of parameters in the encoder layers helps the models to filter the features faster and effectively without causing the overfitting of the model. The effect of self-attention layers in filtering and smoothing the learning curve can also be seen in comparing the loss curve of the CNN and encoder models presented in Fig. \ref{fig:Loss transformer}.

We also tried transfer learning by using an already trained CNN as the backbone of the encoder model. Surprisingly, however,  the encoder models that do not use transfer learning performed slightly better than the models that use transfer learning. Since a trained CNN model has already learned to extract specific features of an image, the encoder model with that CNN backbone is restricted to minimise the loss function in only a small part of the hyperspace. So, the self-attention layers can only filter the features and improve the accuracy by a small percentage (e.g. CNN 2  86.74\% and Lens Detector 2, with CNN 2 as the backbone,  88.13\%). Nevertheless, for a model without transfer learning, there is a possibility for the CNN part in the encoder model to learn more features than a solo CNN about the image and improve the accuracy. For example,  Lens Detector 7, which used the trained CNN 3 as its backbone, scored an accuracy of 91.45\%. In contrast,   Lens Detector 15, which used a CNN backbone similar to CNN 3 and without any transfer learning, scored an accuracy of  92.99\%. However, this result cannot be generalised since it also depends on the trained CNN backbone. 

An updated version of the CNN models that participated in the challenge and  had better scores
in every category compared to their previous versions has recently been reported by \citet{magro2021comparative}.  They used the same CNNs that participated in the challenge and retrained the networks with different epochs. Even though the models had improved  scores, it was evident from the report that the performance of the models is highly dependent on the number of epochs. In other words, the CNNs reported in the Bologna Lens Challenge had lower stability. We have to monitor the training to achieve better results carefully. In contrast, the encoder models  are highly stable compared to the CNNs. As shown in Fig. \ref{fig:Loss transformer}, we were able to train the encoder models up to 2000 epochs without any sign of  overfitting. Interestingly, the fluctuations in the validation loss were very stable up to the end.

\subsection{Lens detectors for strong lens detection}

\begin{figure}[h]
    \centering
\includegraphics[width=250 pt,keepaspectratio]{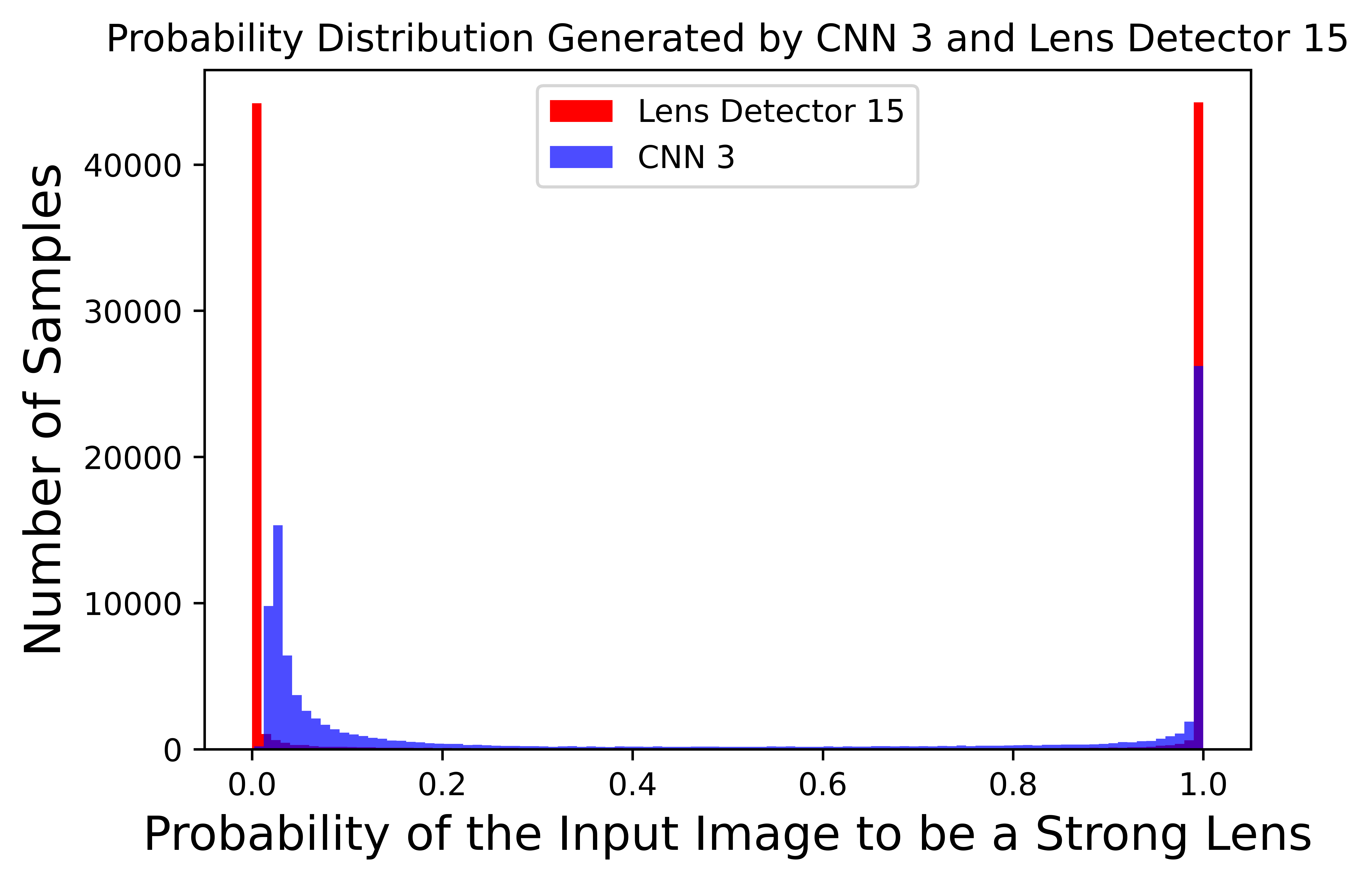}
     \caption{Comparison of the output probabilities of CNN 3 and Lens Detector 15 for the ground-based challenge dataset. In this histogram values leaning towards zero represent the lack of a strong lens in the image, and values leaning towards one indicate the presence of a strong lens in the image.}
    \label{fig:Prediction} 
\end{figure}

It is worth pointing out that  encoder models can identify SLs and non-SLs better than their CNN counterparts. The probability distribution of finding a lens in the challenge dataset is depicted in Fig. \ref{fig:Prediction}. The encoder model can assign a probability for an input to be a lens ($p\approx 1$) or non-lens ($p\approx 0$) with greater confidence than the CNN. Furthermore, from Fig. \ref{fig:Prediction}, it is clear that the transformer models can approximately mimic a perfect classifier by assigning a probability of 0 to non-lensed images and a probability of 1 to lensed images. This feature of the encoder model will be beneficial and applicable in the upcoming large-scale surveys to narrow down the potential lensing systems with great confidence.  Figure \ref{fig:CM15} shows that  Lens Detector 15 was able to identify 82\% of the true lenses with a probability greater than 0.999, which explains the peak near the point 1.0 on the x-axis of Fig. \ref{fig:Prediction}.

\begin{figure}[h]
    \centering
\includegraphics[width=250 pt,keepaspectratio]{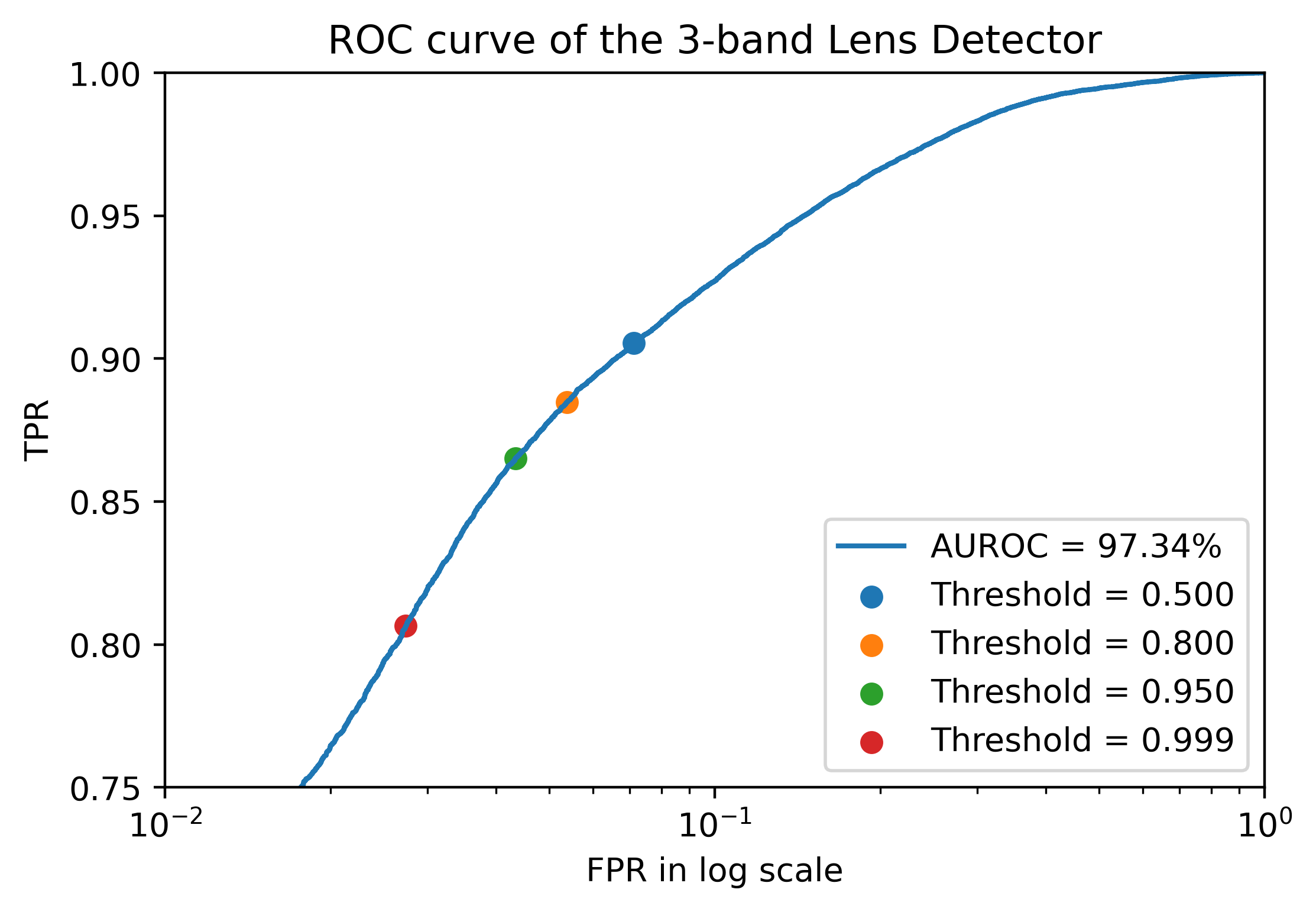}
     \caption{ROC curve of the 3-band Lens Detector.}
    \label{fig:3bandroc} 
\end{figure}

\begin{figure*}
    \centering
\includegraphics[width=500 pt,keepaspectratio]{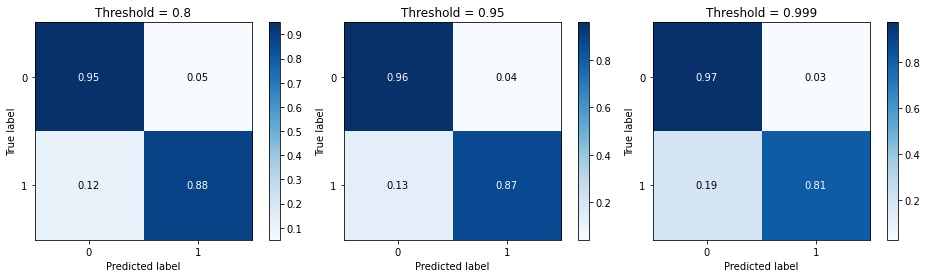}
     \caption{Confusion matrix of the 3-band Lens Detector on the challenge dataset for various thresholds. Class 0 represents the non-lensed images, and Class 1 represents the lensed images. The lower right represents TP in each confusion matrix, the lower left represents FN, the upper left represents TN, and the upper right represents FP.}
    \label{fig:3bands} 
\end{figure*}

Here, we  used all four bands available for training the models. However, the $u$-band  images are usually not used to search for strong lenses because, in the real scenario, the $u$-band images  often have lower image quality. It is also possible that they are not   available for fainter galaxies.
In the literature, for detecting the strong lenses, the images from the $g, r,$ and $i$ bands are used for training machine learning models. Sometimes along with a three-channel CNN, another single-band CNN is also created, and the combined predictions of these two CNNs are used to shortlist the real lens candidates \citep{Petrillo_2018,Petrillo_2019, Li_2020}. To test the adaptability of the encoder model for three bands, we removed the $u$ band and retrained the  Lens Detector 15 model from scratch with the images from the $g, r$, and $i$ bands and named it the 3-band Lens Detector.  The retrained Lens Detector 15 achieved an AUROC = 0.974, which is comparable to the AUROC when the $u$ band was present. The ROC curve and the confusion matrix for various thresholds are presented in Fig. \ref{fig:3bandroc} and Fig. \ref{fig:3bands}. Comparing the confusion matrices in  Fig. \ref{fig:CM15} and  Fig. \ref{fig:3bands}, we can see that removing the $u$ band slightly reduces the number of true positives for a given threshold since the information available in the $u$ band was gone. For example, for a threshold = 0.8,  Lens Detector 15 identified 90\% of the true positives and the 3-band Lens Detector identified 88\% of the true positives. This result also validates the argument presented in \citet{Metcalf_2019} to include even low-resolution information in one band to improve the detection rates. 

\begin{figure}[h]
    \centering
\includegraphics[width=250 pt,keepaspectratio]{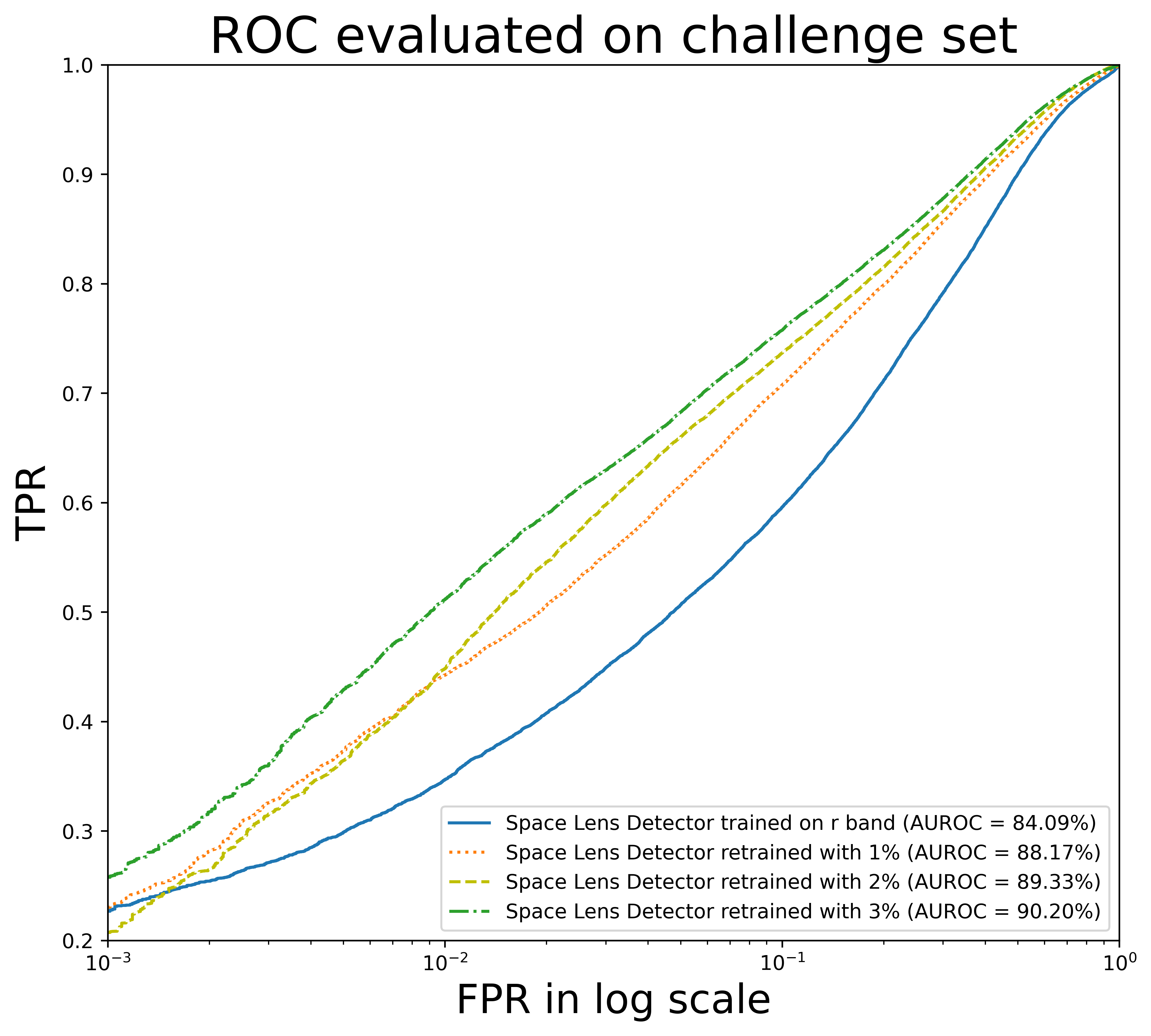}
     \caption{ROC curve of Space Lens Detector trained on $r$ band and tested on the space-based dataset. Improved ROC curves of the model retrained using 200, 400, and 600 samples from the space dataset are also plotted. }
    \label{fig:ROC_S_R} 
\end{figure}

Since we  already tested the encoder model on the space-based dataset and obtained comparable results with models that participated in the Bologna Lens Challenge, it is clear that encoder models can be adapted for a single-band analysis. However, another interesting question to be investigated is the ability of the encoder models to find strong lenses from a different data distribution than the model has been trained on. We retrained the Space Lens Detector using the data in the $r$ band from the ground-based data and tested it on the space-based challenge dataset to investigate this aspect. The retrained Space Lens Detector scored   AUROC = 0.84, which shows the model has the minimum capacity to distinguish lenses from a different distribution. If we train the network again with 200 samples from the space-based dataset (1\% of the space training set), the AUROC improves by 0.88. With 400 samples (2\%), AUROC becomes 0.89, and with 600 samples (3\%), AUROC improves to \textbf{0.902}. Each retraining was done independently of the others. The ROC curve of the Space Lens Detector trained with $r$ band and was tested on the space-based dataset, and the improved ROC curves of the retrained model are shown in Fig. \ref{fig:ROC_S_R}. The capacity of the model to identify the lenses improves if we train the model with very few samples from a different distribution, which indicates the adaptability of the encoder model in the presence of new data distributions. This feature also shows that the encoder models trained on simulated data can be optimised to detect strong lenses from real data using even a small sample of detected lenses.

 \begin{figure*}
    \centering
\includegraphics[width=500 pt,keepaspectratio]{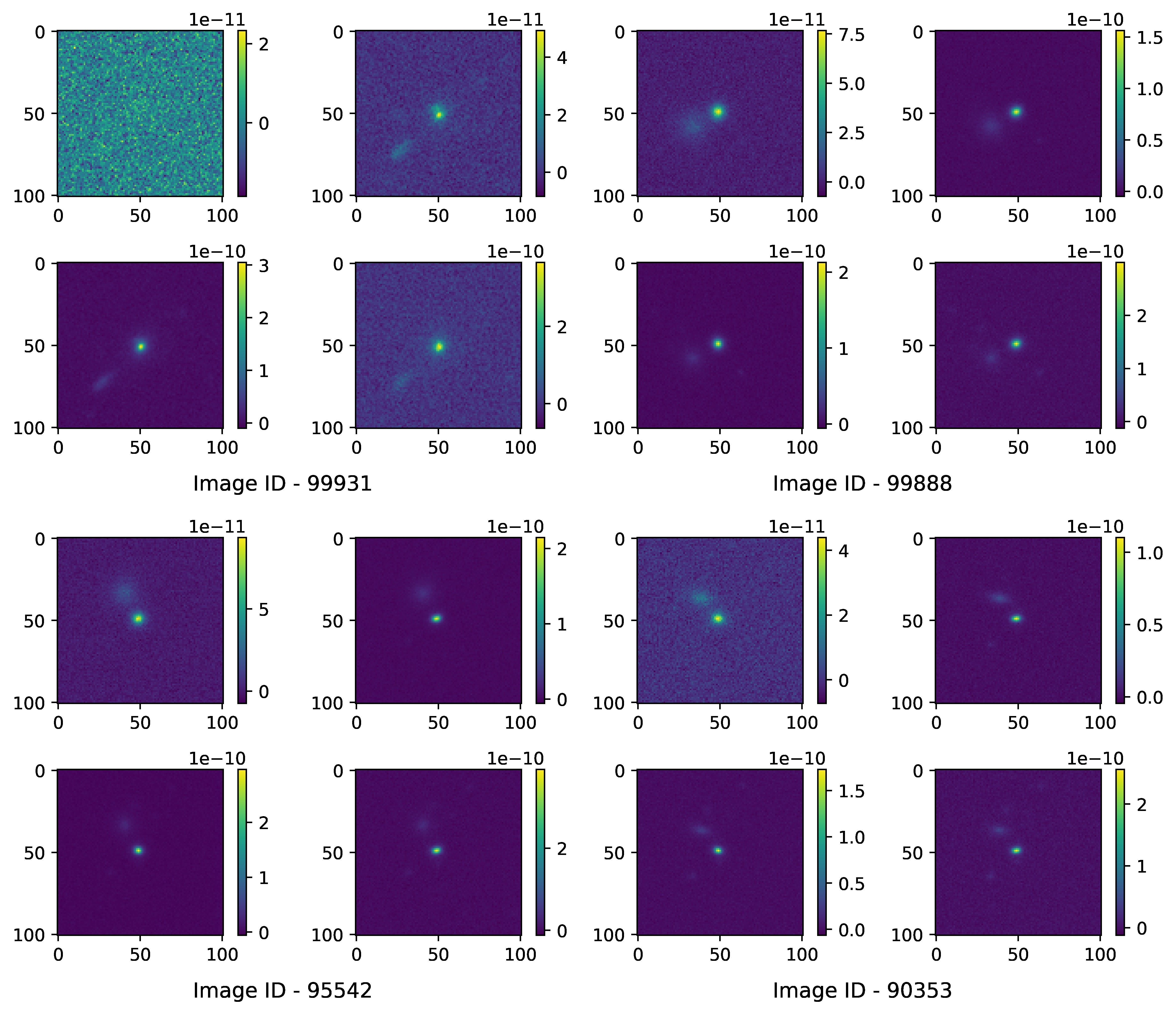} 
     \caption{Four examples of false positives found by the encoder models. The channels  shown are  $u$ (top left), $g$ (top right), $r$ (bottom left), and $i$ (bottom right). Image ID from the test data is given below  each set of images.}
    \label{fig:fp} 
\end{figure*}

\begin{figure*}
    \centering
\includegraphics[width=500 pt,keepaspectratio]{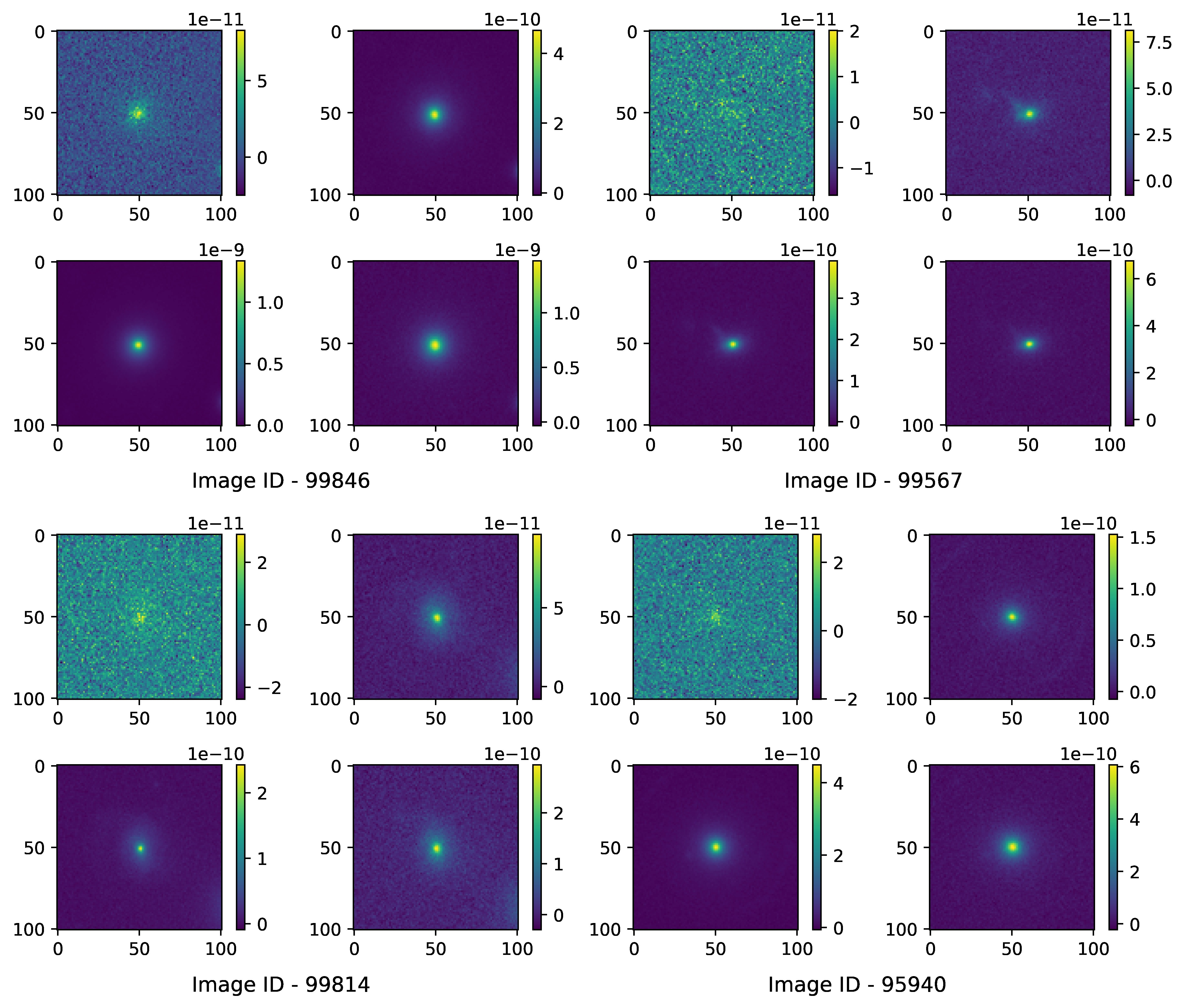} 
     \caption{Four examples of false negatives found by the encoder models. The channels shown are  $u$ (top left), $g$ (top right), $r$ (bottom left), and $i$ (bottom right). Image ID from the test data is given below   each set of images.}
    \label{fig:fn} 
\end{figure*}

Even though the encoder model performs better than the convolution models and the other models that participated in the challenge, the encoder models that have been trained here have a slight gap with a perfect classifier. We carefully examined the frequent false positives and the false negatives reported by various encoder models. Some of the images that have been identified as false positives and false negatives are given in Fig. \ref{fig:fp} and Fig. \ref{fig:fn}. Looking at these false positives and false negatives, we can see that the encoder models are trying to find whether the input image has an arc-like structure or multiple distorted images. If we suppose the input image has any of these characteristics in at least one of the bands,  the detector identifies the image as a strong lens. Similarly, if both these features are missing, then the detector classifies the image as a   non-lens. In order to improve the performance of the models, we need the model to be trained on more realistic and complex data.

\begin{figure*}
    \centering
\includegraphics[width=500 pt,keepaspectratio]{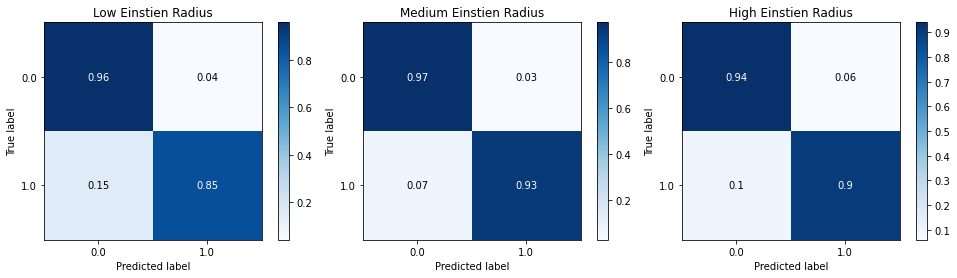} 
     \caption{Confusion matrix of Lens Detector 15 with 0.8 as the threshold plotted for small Einstein radius ($0.3011$ -- $0.873 \text{ arcsec}$), medium Einstein radius ($0.873$ -- $3.547 \text{ arcsec} $) and high Einstein radius ($3.547$ -- $10.08 \text{ arcsec}$). In the confusion matrix, the lower right represents TP, the lower left represents FN, the upper left represents TN, and the upper right represents FP.}
    \label{fig:ER} 
\end{figure*}

Since strong gravitational lensing is a rare phenomenon, we need to have a closer look at the cases of false negatives. In real surveys false positives can be filtered out from a candidate sample created by an automated classifier, whereas missing a true SL is not favoured. Looking at the relations between the parameters of the strong lens and the model's performance is a possible way to search for a space of parameters where it is more difficult to find lenses. We sorted the test dataset into ascending order in Einstein radii and divided it into three subclasses depending on the Einstein radius. The first quartile (25,000) and the third quartile (75,000) were used as the subclass boundaries. Looking at the confusion matrices (threshold = 0.8) from Fig. \ref{fig:ER}, we can see that the best performance of Lens Detector 15 is for the intermediate bin (0.873--3.547 arcsec). The performance of Lens Detector 15 is low for the first bin (0.3011--0.873 arcsec) and the third bin (3.547--10.08 arcsec). This means that Lens Detector 15 has difficulty finding SLs with small and large Einstein radii. A similar result for CNNs has also been reported by \citet{Li_2020,Lens_CNN&A...644A.163C}.

\begin{figure}[h]
    \centering
\includegraphics[width=250 pt,keepaspectratio]{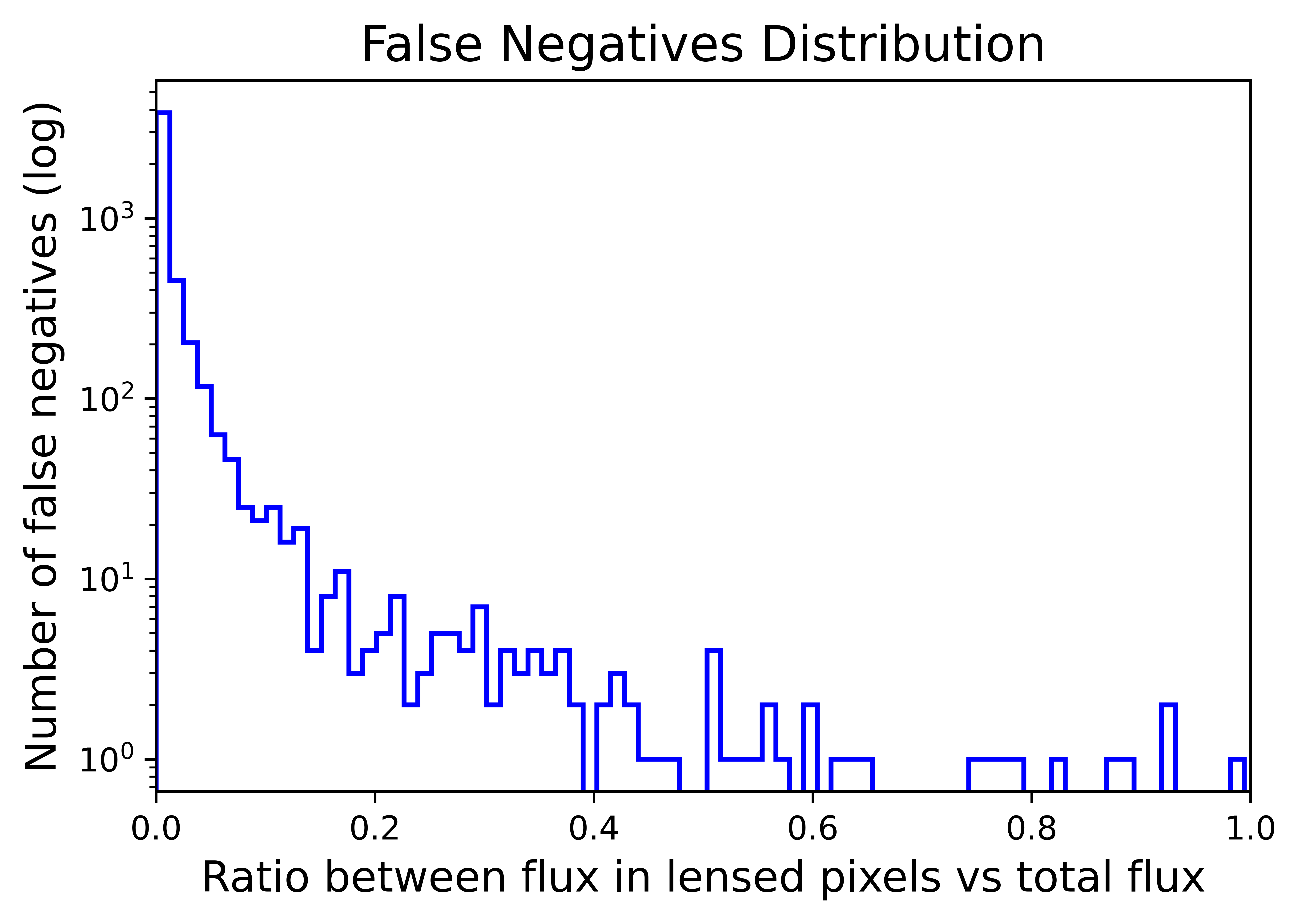} 
     \caption{Distribution of false negatives as a function of the ratio of flux in the lensed pixels to the total flux.}
    \label{fig:LS_ratio} 
\end{figure}

Analysing the parameters of the false negatives, we found that another important parameter that impacts the identification of strong lenses is the ratio of the flux in lensed pixels to the total flux. For a probability threshold of 0.8, we had 4981 false negatives in the Bologna Ground-Based Challenge. All of them had a very low ratio of the flux in lensed pixels to the total flux. Out of 4981 false negative cases, 4775 cases had a flux ratio of the source to the lens lower than 0.1. Similarly, 3667 out of 4981 cases in the sample of false negatives had a flux ratio of the source to the lens  lower than 0.01. The Bologna Ground-Based Challenge dataset contained 10,818 true strong lenses with a flux ratio of the source to the lens lower than 0.01. Thus, Lens Detector 15 could identify 66\% of the true strong lenses with a very low flux ratio (lower than 0.01). These results indicate that  Lens Detector 15 will have trouble distinguishing strong lens candidates for a very low flux ratio between the source and the lens. The reason is that the distortions of the source galaxy due to lensing may not be bright enough to be detected by the models. This result is similar to the results reported by \citet{Li_2020,Lens_CNN&A...644A.163C} where the CNN models have low performance on fainter SLs samples. The distribution of false negatives as a function of the ratio of flux in the lensed pixels to the total flux is plotted in Fig. \ref{fig:LS_ratio}.

We also would like to comment on another reported CNN model on the Bologna Lens Challenge, which is the LensCNN, achieving a total accuracy of 0.8749 (TP  0.8817 and TN 0.8682). It is the only  CNN model where classification accuracy for the Bologna Lens challenge has been reported \citep{2019MNRAS.487.5263D}. Our results show  that all of our encoder models have surpassed the LensCNN in total accuracy. Furthermore, the LensCNN model has also reported an AUROC of 0.96 on the challenge data, which is exceeded by most of our encoder models. In this context, it is worth mentioning that the LensCNN had approximately  10 $\times 10^6$ parameters. Lens Detectors outperformed the LensCNN with just 3 $\times 10^6$ of the parameters.

\subsection{Performance on real data} 

Since all our models have been trained on the simulated dataset provided by the Bologna Lens Challenge, it is critical to check if the trained model can identify strong lenses from real data. Ideally, we expect the encoder models to learn the general features of the strong lenses  from the simulated data and to retrieve the potential lens candidates from the real data. Recently, \cite{Petrillo_2019} have trained a three-band CNN $(g,r$, and $i$ bands) and a single-band CNN  ($r$ band)  on the data simulated using the real images from the KiDS survey and applied it to the KiDS DR4 data to identify potential strong lens candidates. To obtain a reasonable number of true positives and so as not to contaminate the sample with a large number of false positives, they chose 0.8 as the threshold for identifying a candidate as a strong lens for each CNN. Using these criteria, they shortlisted 3500 cases as potential strong lenses, and     \citet{Petrillo_2019} conducted a visual inspection to validate these candidates.

The potential candidates were classified into three classes,  and each class was assigned a numerical score: Sure lens, 10 points; Maybe lens, 4 points; No lens, 0 points. 
As a result, the highest score that any candidate can obtain is 70, when all human classifiers think it is undoubtedly a lens. Using these criteria, \citet{Petrillo_2019} shortlisted 1983 potential strong lens candidates from the data selected by the two CNNs. The FITS files, probability scores of the CNNs reported in \citet{Petrillo_2019}, and numerical scores of visual inspection for each candidate are available publicly  \footnote{\hyperlink{https://www.astro.rug.nl/lensesinkids/}{\text{https://www.astro.rug.nl/lensesinkids/}}}, and we chose this dataset to study the performance of the encoder model on real data.

\begin{figure*}
    \centering
\includegraphics[width=450 pt,keepaspectratio]{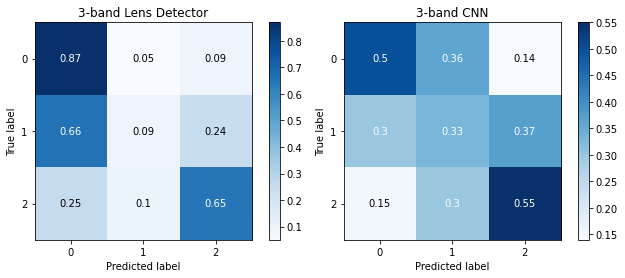} 
     \caption{Confusion matrix of the 3-band Lens Detector and three-band CNN for the classification of the 1983 KiDS DR4 lens candidates \citep{Petrillo_2019}. The candidates are classified into three categories based on the visual inspection scores by \citet{Petrillo_2019}, which is treated as the correct classification. The candidates are also classified into three categories based on the probability values generated by the 3-band Lens Detector introduced in this work and three-band CNN from \citet{Petrillo_2019}. The overlap between these classifications is shown in the confusion matrices.} 
    \label{fig:real_data} 
\end{figure*}

 Since the 1983 lens candidates were found together using a single-band CNN and a three-band CNN, some of the candidates found by the single-band CNN were not detected by the three-band CNN. Specifically, 946 candidates were missed by the three-band CNN (which means they were assigned a probability of less than 0.8) in the final sample of 1983 candidates. Similarly, 526 candidates identified by the three-band CNN were missed by the single-band CNN. 
To analyse the performance of the encoder model, we used the 3-band Lens Detector and tested it on the lens candidates found by \citet{Petrillo_2019}. After evaluating the model on real data, we created three classes using the visual inspection score as the reference.
Class 0 sources have a  low probability of being a lens (score less than 20 out of 70, and predictive value less than 0.8).
Class 1 sources have an  intermediate probability of being a  lens (score between 20 and 50, and predictive value between 0.8 and 0.95). 
We are highly confident that Class 2  sources   are  strong lenses (score greater than 50, and a predictive value greater than 0.95). Using the probability scores predicted by the three-band CNN and using the visual scores, we plotted the confusion matrix for the three-band CNN along with the 3-band Lens Detector, which is given in Fig. \ref{fig:real_data}.

Looking at the confusion matrix in Fig. \ref{fig:real_data}, we can see that the encoder model can classify low (Class 0) and high (Class 2) probability cases similarly to a human expert. However, for class 1, which represents the cases with an intermediate probability of being a strong lens, the 3-band Lens Detector performs poorly. This is to be expected since the 3-band Lens Detector is trained as a binary classifier, and from Fig. \ref{fig:Prediction} it is clear that the Lens Detector tends to assign very high or very low probability scores. Since there are three classes and the number of samples in each class is different, in order to compare the three-band CNN  and the 3-band Lens Detector, we can calculate the weighted  $f_1$ score by taking the mean of all per class $f_1$ scores while considering each class's support. Here, support refers to the number of actual occurrences of the class in the dataset. Using the visual scores as a reference, the weighted $f_1$ score of the  three-band CNN is 0.601, and the weighted $f_1$ score of the 3-band Lens Detector is 0.822, which indicates that the 3-band Lens Detector is performing similarly to a human visual expert on the shortlisted SL candidates. Here we have assumed that the output probabilities assigned by the three-band CNN and the team of visual experts are independent.

With these results, we cannot claim that the 3-band Lens Detector is better than the three-band CNN presented in \citet{Petrillo_2019} or vice versa since we are testing the model on the already shortlisted candidates by the three-band CNN. However, we can claim that self-attention-based encoder models can detect strong lenses from the real data in competition with the CNNs. Another factor to be noted here is that the 3-band Lens Detector was trained on a complete data distribution compared to the training set of the three-band CNN, which was derived from the actual KiDS DR4 data. As mentioned earlier, the data in the Bologna Lens Challenge used KiDS as a reference, and they did not strictly mimic the data from  KiDS. Thus, the data used in the Bologna Lens Challenge have a different data distribution compared to the KiDS DR4 data. Above  we  showed that the encoder models can adapt to different data distributions and improve their performance if we retrain, even with a small sample set from the new data distribution. Thus, the performance of the encoder model can be significantly improved if one retrains the 3-band Lens Detector with the data derived from the KiDS DR4 data.

\section{Conclusions}

We have presented a novel machine learning approach known as the self-attention-based encoders to detect strong gravitational lenses. We have explored this new architecture's possibilities to understand better how to apply the transformer models for image analysis using the data from the Bologna Lens challenge. Currently, most of the automated techniques employed to find strong lenses are based on CNNs. However, as noted by \citet{Metcalf_2019}, CNNs are prone to overfitting the training set. Here we showed that the self-attention-based architectures provide better stability and are less likely to overfit than CNNs. Another  advantage of a self-attention-based encoder over a CNN is that it performs better with a fewer number of trainable parameters. Hence, self-attention-based encoder models can be considered a better alternative to CNNs and other automated methods.   

Here we have described the   21 encoder models we created to study the application of self-attention-based models for SL detection using the data from the Bologna  Lens Challenge. We have presented the three best encoder models, which  provide    more reliable performance than those  participating in the Bologna  Lens Challenge.
 Lens Detector 21 scored an AUROC of \textbf{0.9809}, which is equivalent to the top AUROC achieved in the challenge. Similarly,  Lens Detector 16 scored a $TPR_0$ \textbf{0.225} higher than any model that participated in the challenge, and surpassed the top TPR$_0$ ($0.09$) achieved by the CNNs by a high margin. We consider  Lens Detector 15 to be  the best encoder model as it scored 0.14 and 0.48 respectively for $TPR_0$ and $TPR_{10}$, outperforming the CNN models to a greater extent and also scoring an AUROC of 0.9783, which is very close to the top AUROC. 

From our analysis, we were able to point out that the encoder models have more stability than CNNs, which minimises the need for human interaction or monitoring. Similarly,  the encoder models were better than the CNN models in classifying lenses and non-lenses by assigning high probability scores for the lens ($p\approx 1$) and  non-lens ($p\approx 0$) systems.
In addition, the architecture we proposed here is very simple and robust and has a high resistance to overfitting. We could train models as deep as 25 layers and for 2000 epochs without any sign of overfitting. With a simple 8-layer deep CNN, we were able to surpass the performance of a 46-layer deep RNN and surpass all the other models to a great extent.

We tested our model on the 1983 potential strong lens candidates from the KiDS DR4 data found in \citet{Petrillo_2019}. We were able to closely mimic a human visual expert in identifying the strong lenses. Even though we cannot claim to outperform the CNN model presented in \citet{Petrillo_2019}, we   confirm that the encoder models can perform well on the real data. Since we have tested the network on a different data distribution than it was trained on, we expect to improve the performance of the encoder model if the training and testing data distribution is similar. In the future we are planning to train the encoder models on more complex data derived from the real data and test them on the real data to find more potential lens candidates. Even though we have glimpsed at the adaptability of the encoder models to different data distributions, further studies are needed to establish the full scope of this architecture.

In the upcoming era of big data in astronomy, automated methods are expected to play a crucial role. Better and alternative automated methods have to be consistently investigated to advance the scientific study in this scenario. From our study it is clear that the search for strong lenses in the current and upcoming wide-field surveys such as KiDS \citep{2019A&A...625A...2K}, HSC \citep{2019PASJ...71..114A}, DES \citep{2021}, LSST \citep{Ivezi__2019,verma2019strong}, Euclid \citep{scaramella2021euclid}, and WFIRST \citep{koekemoer2019ultra} can be achieved using self-attention-based encoder models with better performance compared to CNNs. 

\begin{acknowledgements}
We are thankful to the Referee for their helpful comments, which allowed us to improve the paper significantly. Authors and NCNR are grateful for financial support from MNiSW grant DIR/WK/2018/12 and NCN grants UMO-2017/26/M/ST9/00978 and UMO-2018/30/M/ST9/00757.
\end{acknowledgements}

\bibliographystyle{aa}
\bibliography{Revised}

%
%
\begin{appendix}
\section{ROC of encoder models }
\begin{figure}[H]
    \centering
    \includegraphics[width=500 pt,keepaspectratio]{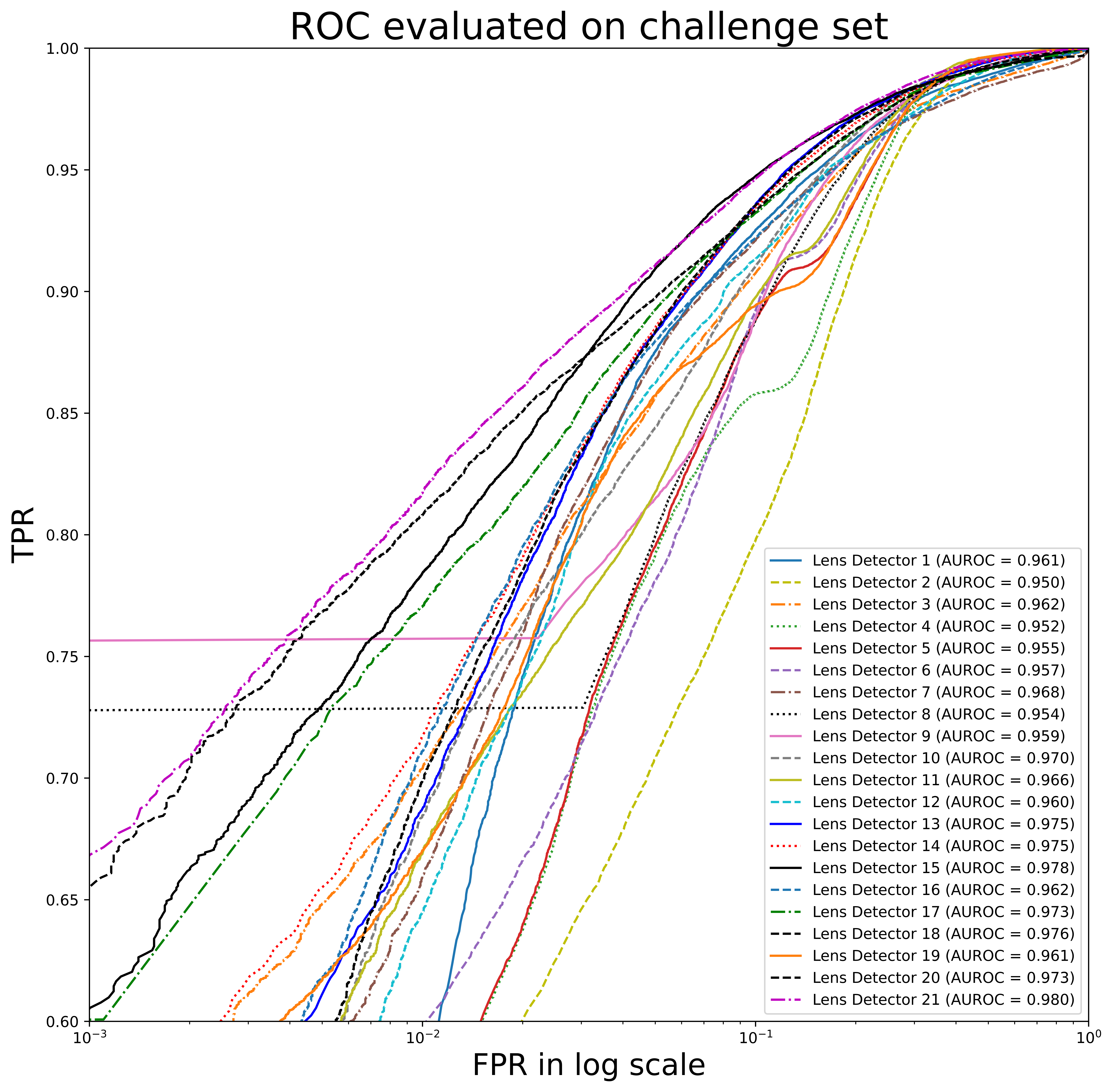}
    \caption{ROC curves of all the encoder models.}
    \label{fig:my_label}
\end{figure}

ROC curves of all the encoder models presented in Table \ref{table:1} are displayed in Fig. \ref{fig:my_label}. The inset gives the AUROC of each model in order to facilitate comparison of the models.
\end{appendix}
\end{document}